\documentclass[10pt,twocolumn,letterpaper]{article}

\usepackage[pagenumbers]{cvpr}

\newcommand{\blue}[1]{{\color{blue}#1}}

\newcommand{\TODO}[1]{\textbf{\color{red}[TODO: #1]}}
\renewcommand{\blue}[1]{#1}  %

\usepackage{microtype}
\usepackage{tikz}
\usetikzlibrary{arrows.meta,positioning,fit,backgrounds,shapes.geometric}
\usepackage{multirow}
\newcommand{\best}[1]{\textbf{#1}}
\newcommand{\secbest}[1]{\underline{#1}}
\usepackage{array}
\usepackage{xspace}
\usepackage{soul}
\usepackage[section]{placeins}  %

\newcommand{\paper}{ProtoDiffusion\xspace}
\newcommand{\protovae}{ProtoVAE\xspace}
\newcommand{\qil}{Qwen-Image-Layered\xspace}

\renewcommand{\TODO}[1]{}

\renewcommand{\blue}[1]{#1}

\definecolor{cvprblue}{rgb}{0.21,0.49,0.74}
\usepackage[pagebackref,breaklinks,colorlinks,allcolors=cvprblue]{hyperref}

\def\confName{CVPR}
\def\confYear{2027}

\title{ProtoDiffusion: Token-per-Layer Generative Modeling of Multi-Layer Images}

\author{%
Adri\`a Ciurana Lanau\textsuperscript{1}\quad
Harleen Hanspal\textsuperscript{1}\quad
Lusine Abrahamyan\textsuperscript{1}\quad
Emilio Tylson Baixauli\textsuperscript{1}\\
Yu Cao\textsuperscript{2}\quad
Danai Triantafyllidou\textsuperscript{1}\quad
Stefano Zorzi\textsuperscript{1}\quad
Gustavo Zanoni Felipe\textsuperscript{3}\\
Peter Griger\textsuperscript{1}\quad
Stefan Sietzen\textsuperscript{1}\quad
Petru-Daniel Tudosiu\textsuperscript{1}\\[5pt]
\textsuperscript{1}Canva\qquad
\textsuperscript{2}Queen Mary University of London\qquad
\textsuperscript{3}Universidade Estadual de Maring\'a\\[3pt]
{\tt\small \{adria,hhanspal,lusine,emilio,danai\}@canva.com}\\
{\tt\small \{szorzi,gzanoni,petergriger,stefan,ptudosiu\}@canva.com}
}

\begin{document}
\raggedbottom

\twocolumn[{%
\renewcommand\twocolumn[1][]{#1}%
\maketitle
\begin{center}
  \centering
  \includegraphics[width=\textwidth]{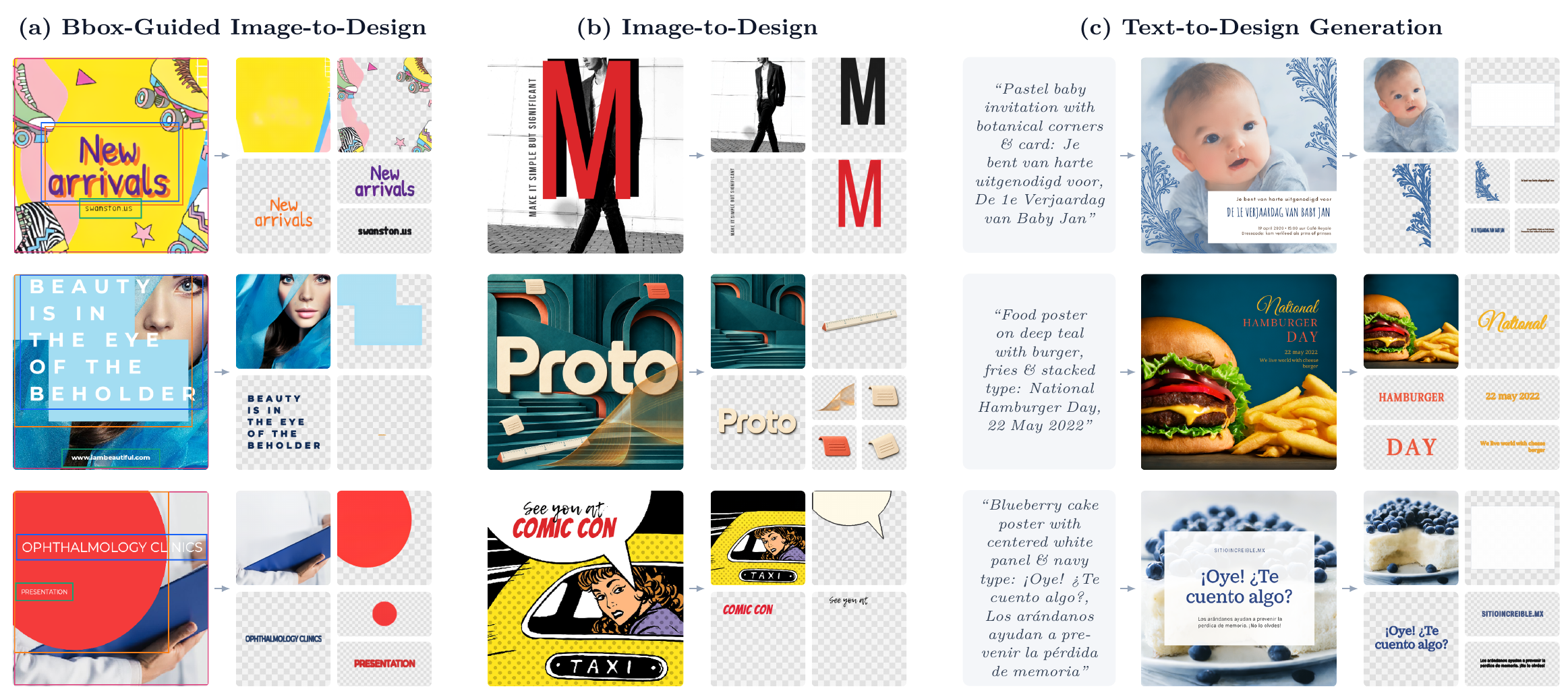}
  \captionof{figure}{\paper{} across its conditioning regimes. (a)~Box-guided
  image-to-design: the input composite, with supplied element boxes, is
  decomposed into editable RGBA layers. (b)~Image-to-design: the same
  decomposition from the composite alone, without boxes. (c)~Text-to-design:
  generation of a layered design from a text prompt. Each group shows the
  composite at left and the recovered layers at right.}
  \label{fig:teaser}
  \vspace{4pt}
\end{center}%
}]

\begin{abstract}
Graphic designs are ordered compositions of RGBA layers whose separation enables selection,
movement, and reuse. Existing layered models represent each layer with a full-canvas or
region-sized spatial latent grid, so representation size grows with the cumulative spatial extent
of the layers. We present \paper, a latent-diffusion framework built on \protovae, a multi-layer
variational autoencoder that encodes a design into one shared spatial prototype and one
candidate vector per layer. Each candidate conditions a shared decoder to reconstruct its
corresponding RGBA layer. For $N$ layers, the representation contains $HW+N$ tokens
before backbone-specific packing, compared with $NHW$ for full-grid-per-layer representations.
\paper jointly denoises the prototype and a fixed set of candidate slots, conditioned on an image or text, with bounding boxes as optional layout guidance.
The model leaves unneeded slots empty and determines the number of layers and, in absence of boxes, their layout, requiring neither as input.
The same architecture supports both image-to-design and text-to-design. ProtoVAE achieves high-fidelity reconstruction from its compact shared latent
representation, reaching 40.43\,dB PSNR on layer crops.
On the LayerD evaluation split, image-to-design achieves RGB L1 scores of 0.040 with bounding boxes
and 0.057 without, with box-guided alpha soft IoU (0.850) comparable to the strongest
fine-tuned baseline (0.871).
On merged images, text-to-design reaches 66.41 FID versus 86.84 for
FLUX.2 [klein], achieves a 0.860 text-accuracy score, and runs $19.1\times$ faster than the
same-backbone generation-then-decomposition cascade.
\end{abstract}

\section{Introduction}
\label{sec:intro}

Graphic designs are ordered compositions of RGBA elements: text, shapes, cutouts, and photographs
over a background. Blending these layers produces an \emph{alpha-composited image}.
Keeping these elements separate enables selection, movement, and reuse after
generation. Standard text-to-image diffusion~\cite{rombach2022ldm} instead produces a
flattened raster, in which those editing units are entangled.

Layered models already generate coherent transparent elements, but they share two constraints.
First, every layer has its own spatial latent: a full-canvas grid in
\qil (QIL)~\cite{qwenimagelayered2025} and a region grid in MRT~\cite{tang2026mrt}.
The sequence processed by the denoiser, and with it compute and memory, therefore grows with
the number and extent of layers (Section~\ref{sec:exp-efficiency}).
Second, structure is decided before content: QIL takes a requested layer count, and MRT a
layout that is supplied or estimated beforehand, so the number of layers, and in MRT their
placement, is fixed before denoising begins.
The two are linked: deciding structure jointly with content means denoising many candidate
layers, which is costly when each one is a spatial grid.

\begin{figure*}[tbp]
\centering
\input{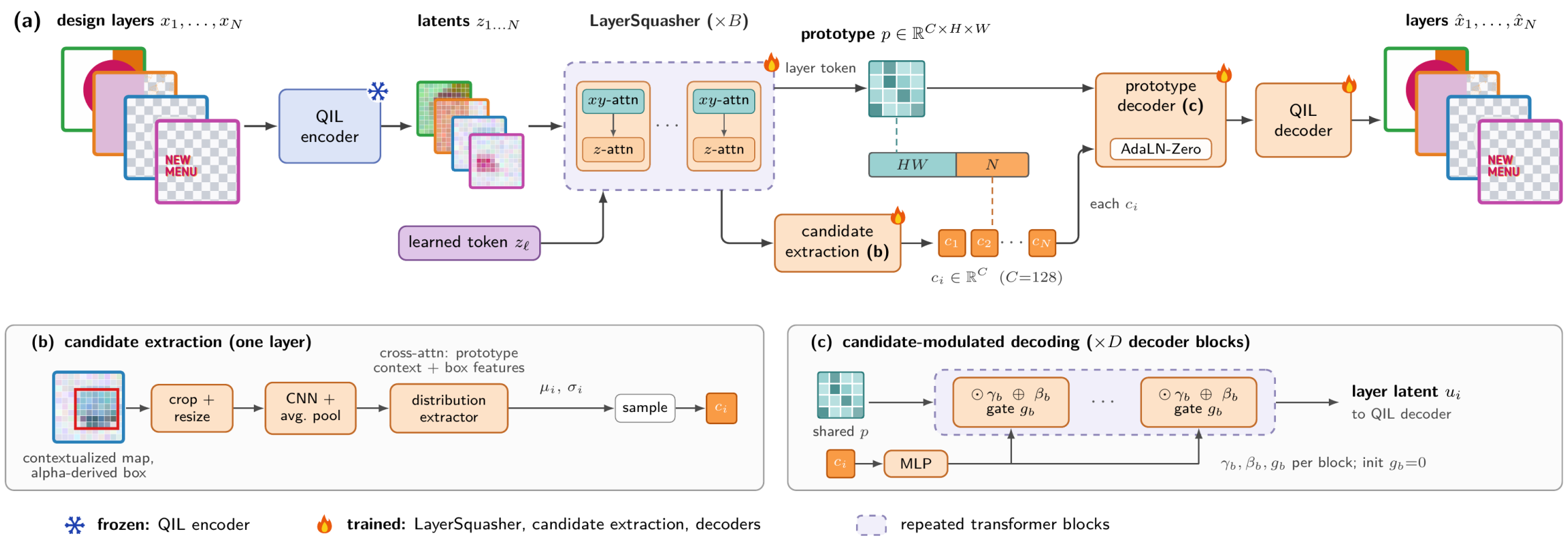}
\caption{\textbf{ProtoVAE: shared grid, one candidate per layer.}
(a) The frozen QIL encoder maps each RGBA layer to a spatial latent; a layer-wise transformer jointly
processes the projected layer maps with a learned layer token, whose output parameterizes the
shared prototype posterior, giving $HW+N$ tokens per design before backbone-specific
packing. (b) Candidate extraction maps each contextualized layer map to a candidate posterior:
a crop around the alpha-derived box, a CNN with average pooling, and a cross-attention
distribution extractor. (c) Each candidate modulates the shared prototype decoder through
AdaLN-Zero, and the jointly trained QIL decoder reconstructs its RGBA layer. Snowflakes mark
frozen modules and flames mark trainable modules. Feature grids and outputs are schematic.}
\label{fig:protovae-architecture}
\end{figure*}

The use of a prior layout is analogous to two-stage instance segmentation
(e.g., Mask R-CNN~\cite{he2017maskrcnn}): propose regions, then predict instances.
As in that setting, the second stage cannot revise what the first provides: a missing or
misplaced region is not corrected during layer generation, and the layout must come from a
separate model or the user.
Inspired by the shared prototype masks and per-instance coefficients in YOLACT~\cite{bolya2019yolact}, 
we instead factorize a design into a shared spatial grid and layer-specific candidates,
extending the idea to joint generation and nonlinear decoding of RGBA layers.
We propose a \emph{token-per-layer} representation.
\protovae{} jointly encodes the layer stack into a shared spatial grid (the \emph{prototype}) and one candidate vector per layer, including
the background. Each candidate summarizes its layer in the context
of the full composition and conditions the decoder to reconstruct
the corresponding RGBA layer from the shared grid
(Figure~\ref{fig:protovae-architecture}).

\paper learns a diffusion model over this factorization. A pretrained image-transformer stream
processes the shared grid, while a dedicated candidate stream exchanges information with it through
bidirectional cross-attention (Figure~\ref{fig:architecture}).
The same model serves different tasks by changing the conditioning: image-to-design conditions on
the encoded alpha-composited image, and text-to-design on a prompt. Boxes guide candidates
when available.
The model denoises a fixed set of candidate slots and leaves unneeded ones empty. It thus
determines the number of layers and, when no boxes are given, their layout, requiring neither as input. 
Figure~\ref{fig:teaser} illustrates image-to-design
with and without boxes and text-to-design generation.
Specifically, our contributions are:
\begin{enumerate}
  \item \textbf{Compact latent representation.} A token-per-layer factorization, \protovae,
  represents an RGBA design with $HW+N$ tokens before packing for $N$ layers, versus
  $N\!\cdot\!HW$ for full-grid-per-layer representations. Each additional layer costs one
  candidate vector rather than a spatial grid. Section~\ref{sec:token-efficiency} derives the
  resulting denoiser scaling separately from measured system-level speedups.
  \item \textbf{Joint denoising of design context and elements.} A two-stream diffusion
  architecture couples the shared prototype grid and layer candidates through bidirectional
  cross-attention, jointly modeling composition and layer content.
  \item \textbf{Unified generation and decomposition.} A single model supports image-to-design
  with or without bounding boxes and text-to-design generation through changes in conditioning.
  Activity-based candidate selection recovers a variable number of layers within a fixed capacity.
\end{enumerate}

\section{Related Work}
\label{sec:related}

\paragraph{Layered image generation and decomposition.}
Layered representations allow individual elements of an image
to be edited and reused.
LayerDiffuse~\cite{zhang2024layerdiffuse} introduces latent
transparency, enabling pretrained diffusion models to generate
transparent images and multiple interacting layers.
\qil~\cite{qwenimagelayered2025} encodes each RGBA layer as a
spatial latent grid and models interactions across layers with
a variable-length diffusion transformer. Its decomposition
interface takes an image and a requested layer count, without
requiring bounding boxes.
MRT~\cite{tang2026mrt} uses region-based spatial latents and a
masked diffusion formulation for text-to-layer generation,
image decomposition, and layer editing. Its layer generator
receives a layout specifying the regions and their ordering.
PSDiffusion~\cite{huang2026psdiffusion} generates multiple
layers simultaneously, with global interactions that
coordinate their layout and appearance.

For image decomposition, LayerD~\cite{suzuki2025layerd}
iteratively extracts unoccluded foreground layers through
matting and completes the remaining background through
inpainting.
RevealLayer~\cite{revealllayer2026} focuses on natural-image
decomposition, including occluded content and layer
separation, and introduces RevealLayerBench.
LayeringDiff~\cite{kang2025layeringdiff} first generates a
composite image and then uses generative priors to disassemble
it into foreground and background layers.
For iterative editing, SLEDGE~\cite{khan2026sledge} studies
sequences of instruction-guided layered edits, while
Wu et al.~\cite{wu2026syntheticlayered} investigate the effect
of synthetic layered training data on design decomposition.

\protovae addresses the spatial cost of these representations by sharing one prototype
across layers and adding only one token per layer, instead of a full or regional grid.

\textbf{Graphic design and layout generation.}
COLE~\cite{jia2023cole} uses hierarchical design planning and generation, while
Desigen~\cite{weng2024desigen} predicts layouts conditioned on generated backgrounds.
LayoutDM~\cite{inoue2023layoutdm} and LayoutGPT~\cite{li2023layoutgpt} generate layouts
through discrete diffusion and language models, respectively.
Our framework accepts optional boxes and otherwise jointly models spatial structure and
layer content from text or image conditioning.

\textbf{Latent generative representations.}
We build on latent diffusion~\cite{rombach2022ldm}, transformer
denoisers~\cite{peebles2023dit}, and flow matching~\cite{lipman2023flowmatching} to
jointly model the shared prototype and per-layer candidates. The supplementary material
discusses object-centric representations and reconstruction objectives.

\section{Method}
\label{sec:method}

\begin{figure*}[tbp]
\centering
\input{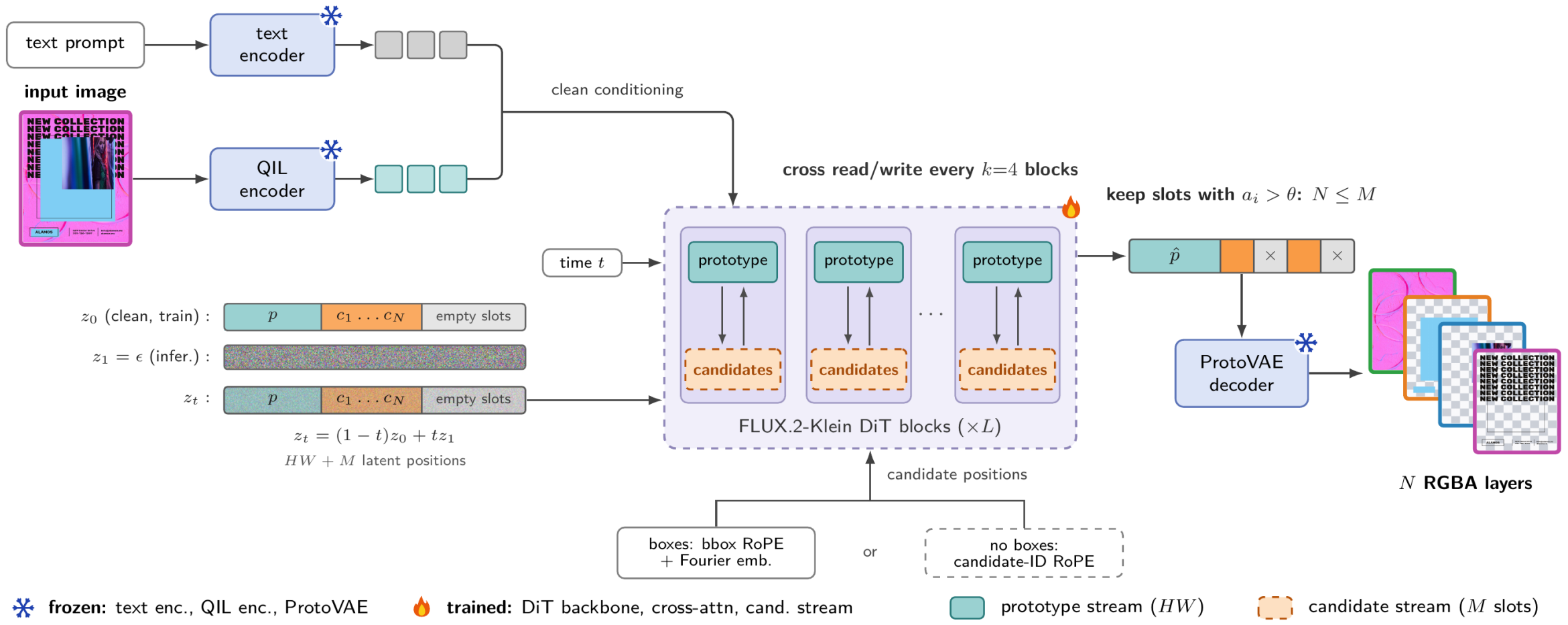}
\caption{\textbf{ProtoDiffusion: conditioning, noising, and joint denoising.}
Image and text conditioning remain clean while the shared prototype and $M$ candidate slots
are jointly denoised through interacting streams. Training interpolates the clean state $z_0$
and Gaussian noise $z_1$ as $z_t=(1-t)z_0+t z_1$; generation starts at $z_1$.
Optional bounding boxes provide spatial guidance. Activity-based selection retains $N\le M$
candidates, which are decoded with the predicted clean prototype $\hat p$ into RGBA layers.
The state contains $HW+M$ tokens before backbone-specific packing. Snowflakes and flames
indicate frozen and trainable modules during diffusion training. Output thumbnails are schematic.}
\label{fig:architecture}
\end{figure*}

We adopt a two-stage training procedure: \protovae learns a latent representation of layered
designs, and \paper is subsequently trained in this latent space.
Figures~\ref{fig:protovae-architecture} and~\ref{fig:architecture} illustrate the representation
and denoising architecture, respectively. Throughout, $H$ and $W$ denote the latent spatial
dimensions, $C$ the number of ProtoVAE latent channels, $N$ the number of design layers
(including the background), and $M$ the number of diffusion slots.

\subsection{\protovae: Shared Spatial Context and Layer Identity}
\label{sec:method-protovae}

\paragraph{Encoding.}
Given an ordered design $\{x_i\}_{i=1}^{N}$ including its background, the frozen QIL VAE
encoder maps each RGBA layer to a spatial latent, which is projected to hidden width $D$.
The projected layer maps are jointly processed by \emph{LayerSquasher}, a layer-wise vision
transformer that alternates spatial self-attention within each layer ($xy$) with self-attention
across layers ($z$). This joint processing provides shared spatial context for constructing
the prototype and per-layer candidates.
A learned token $z_\ell\in\mathbb{R}^{D}$ is broadcast into an additional spatial layer of size
$D\times H\times W$ and jointly processed with the $N$ projected layer maps.
Its output aggregates information from the entire design and parameterizes the shared prototype
posterior. The remaining $N$ outputs are pooled to parameterize candidate posteriors, each
representing a layer in the context of the full design.

Specifically, this pooling operation crops each contextualized layer feature map to a region
around its alpha-derived bounding box and resizes it. A small CNN followed by global average
pooling summarizes the region into a vector. An attention-based distribution extractor refines
this vector using prototype context and box features. The boxes are derived from observed layers
during encoding; they are not externally supplied generation-time layouts.
The resulting diagonal-Gaussian posteriors define the prototype
$p\in\mathbb{R}^{C\times H\times W}$ and candidates $c_i\in\mathbb{R}^{C}$, with $C=128$.

\textbf{Decoding.}
Each candidate conditions the shared prototype decoder through AdaLN-Zero~\cite{peebles2023dit}, producing a spatial
latent for that layer; the QIL decoder converts it to RGBA pixels. During ProtoVAE training,
the QIL encoder remains frozen, while LayerSquasher, the pooling operator, the prototype decoder,
and the QIL decoder are trained jointly. This fine-tuning adapts the QIL decoder
to design layers with small elements, fine details, and large transparent regions;
the adapted decoder is retained for subsequent reconstruction and generation. All candidates therefore
share spatial context, but each selects its own reconstruction. The supplementary material
gives the encoder and decoder details, including candidate refinement and training implementation.

\textbf{Training objectives.}
We train \protovae with pixel- and latent-space reconstruction losses, KL regularization
of the prototype and candidate posteriors, prototype alignment, and adversarial supervision.
Reconstruction supervises complete RGBA layers, including occluded content. Adversarial
supervision uses a discriminator on stacked layer latents and a PatchGAN~\cite{isola2017pix2pix}
on alpha-composited local crops. The supplementary material provides further objective
and training details.

\textbf{Prototype alignment.}
Inspired by DC-Gen~\cite{he2025dcgen}, an alignment module encourages agreement between
features of the shared prototype and the frozen QIL encoding of the composited image.
This regularizes shared spatial context, while per-layer reconstruction supervises hidden
content; alignment does not require the prototype to equal the composite latent.

\textbf{Overlap and leakage weighting.}
We increase pixel- and latent-space reconstruction weights in regions of inter-layer
overlap to emphasize occluded content. Just outside each layer's support, alpha-only
weighting penalizes leakage without constraining invisible RGB.

\subsection{Joint Prototype and Candidate Diffusion}
\label{sec:method-diffusion}

The transformer adapts FLUX.2-Klein-4B~\cite{flux2klein2025} to the ProtoVAE
prototype--candidate latent space. Prototype tokens use the backbone's image self-attention
and spatial RoPE~\cite{su2021roformer}. Candidates have a dedicated self-attention stream
with AdaLN-Zero timestep conditioning. Bidirectional cross-attention synchronizes the streams every
$k=4$ blocks, allowing layer identities and spatial context to interact during denoising.

\textbf{Conditioning and occupancy.}
Training slots follow increasing stacking level ($z$-layer), then top-to-bottom and left-to-right
box-center order within each level. Generated layers are alpha-composited in retained slot order.
For image-to-design, the alpha-composited input image is encoded directly by the frozen QIL encoder,
without running the layered-input ProtoVAE encoder. For text-to-design, the prototype and $M$ candidate
slots are sampled jointly from noise under text conditioning. Candidate conditioning and occupancy are handled as follows:
\begin{itemize}
  \setlength{\itemsep}{0pt}
  \item \textbf{With boxes (layout-guided):} box-derived RoPE and Fourier box embeddings
  provide spatial conditioning.
  \item \textbf{Without boxes:} candidate-ID RoPE identifies the slots, and box embeddings
  are zeroed; no external layout is supplied.
  \item \textbf{Layer count:} unused slots are trained toward an empty state. For each denoised
  candidate vector $\tilde c_i$, we compute $a_i=|\operatorname{mean}(\tilde c_i)|+
  \operatorname{std}(\tilde c_i)$ across channels and retain slots with $a_i>\theta$.
  The highest-scoring slot is always retained, yielding $1\le N\le M$.
\end{itemize}

We set the default candidate capacity to $M=50$.
Occupancy-based selection removes the need to request an exact count, but retains a capacity setting and an occupancy
threshold. Even when boxes specify a count, the current implementation still denoises $M$ slots.
Autoregressive candidate generation with a learned stopping criterion could remove this fixed-slot
constraint; we leave this extension to future work.
The supplementary material reports a threshold-calibration study
on 1,954 samples.

\textbf{Diffusion training.}
We use flow matching with $z_t=(1-t)z_0+t\epsilon$, where
$\epsilon\sim\mathcal{N}(0,I)$, and velocity target $\epsilon-z_0$.
The training objective is
\[
\mathcal{L}=\mathcal{L}_{p}+\lambda_c\mathcal{L}_{c}
+\lambda_{\mathrm{layer}}\bigl(\mathcal{L}_{\mathrm{QIL\text{-}L1}}
+\lambda_{\mathrm{sep}}\mathcal{L}_{\mathrm{sep}}\bigr).
\]
Here, $\mathcal{L}_{p}$ and $\mathcal{L}_{c}$ are mean squared velocity errors,
normalized separately within the prototype and candidate streams.
Real and empty candidate slots are also reduced separately, so candidate supervision
does not scale with the number of padding slots.
For layered supervision, the predicted clean prototype and candidates are decoded through
the frozen ProtoVAE decoder into per-layer QIL latents, retaining gradients with respect
to the predicted prototype and candidates.
The reconstruction term $\mathcal{L}_{\mathrm{QIL\text{-}L1}}$ measures weighted $L_1$
error against the target latents, emphasizing occluded regions and excluding padding layers.
Inspired by RevealLayer~\cite{revealllayer2026}, $\mathcal{L}_{\mathrm{sep}}$ contrasts
each predicted layer with its own target and other layers. The comparison uses corresponding
canvas positions, without pooling or crop recentering.
The supplementary material provides the loss definitions, weighting schemes,
and reduction details.

\subsection{Representation Cost}
\label{sec:token-efficiency}

Before backbone-specific packing, a design's latent representation has
\[
 L_{\mathrm{ProtoVAE}}=HW+N,\qquad
 L_{\mathrm{diffusion}}=HW+M.
\]
A full-grid-per-layer representation has $NHW$ tokens; region grids instead incur an
area-dependent cost. After packing, let $S$ be the spatial token count per grid. At fixed depth,
hidden widths, and conditioning length, our two-stream attention costs
$O(S^2+M^2+SM)$, including cross-stream exchange, versus $O((NS)^2)$ for joint dense attention
over $N$ full grids. Token-wise projections and MLPs process $S+M$ rather than $NS$ tokens.
Thus increasing layer capacity adds candidate tokens instead of spatial grids. With $M$ fixed,
retaining more layers does not lengthen denoising; RGBA decoding still scales with output count.

Figure~\ref{fig:measured-layer-scaling} complements this analysis with system-level latency and
memory measurements, which also reflect differences in backbone, conditioning, and execution.
The supplementary material provides the full complexity analysis.

\section{Experiments}
\label{sec:experiments}

We evaluate reconstruction fidelity, image-to-design decomposition, text-to-design generation,
and computational efficiency.

\begin{figure*}[tbp]
\centering
\includegraphics[width=0.72\textwidth]{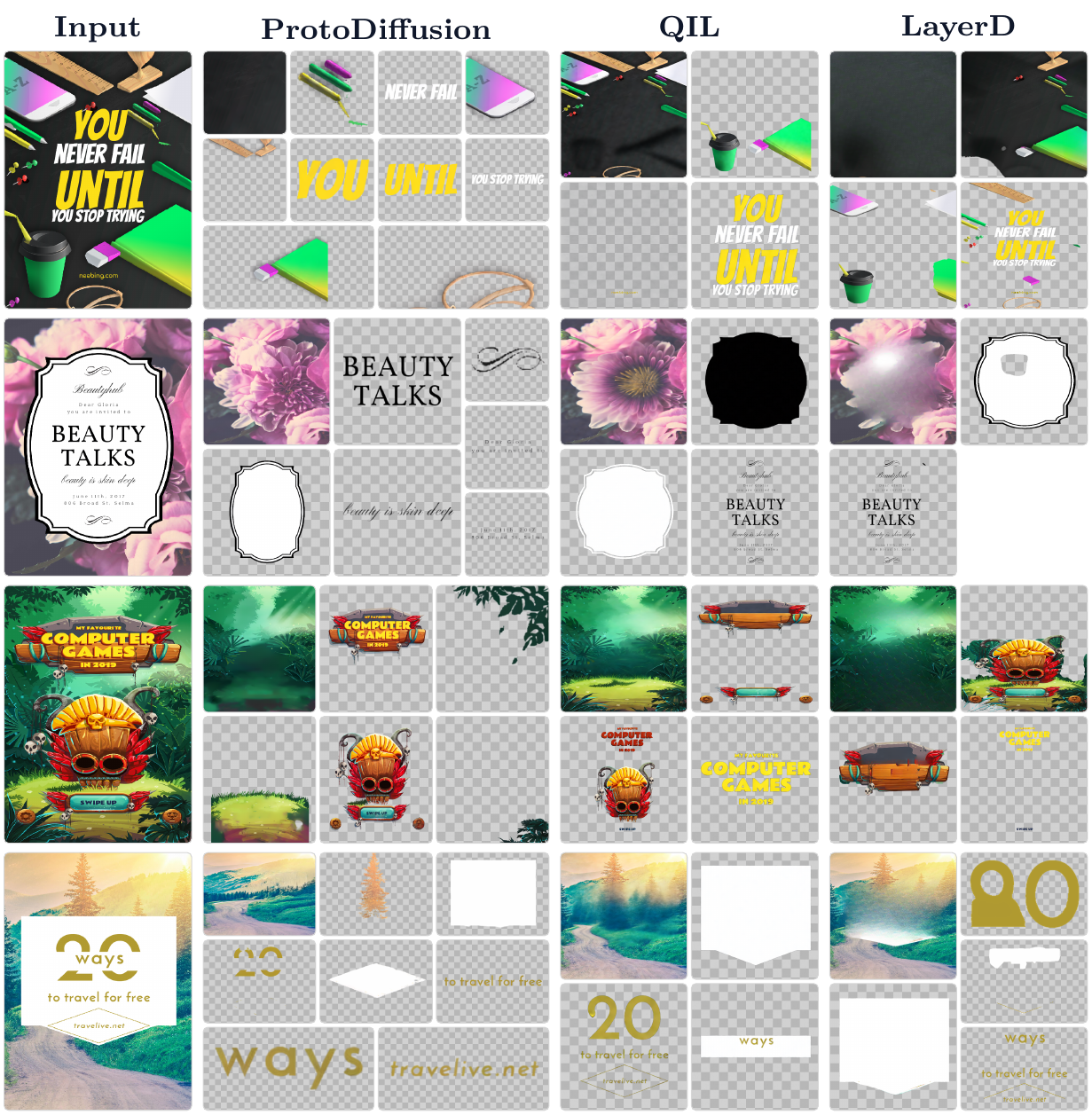}
\caption{Qualitative image-to-design decomposition \emph{without} bounding
boxes on Crello inputs. Columns: the input composite and the layers
recovered by \paper{}, QIL, and LayerD, shown over a checkerboard.
Examples are selected and illustrative, not a systematic comparison.}
\label{fig:comp-nobbox}
\end{figure*}

\begin{figure*}[tbp]
\centering
\includegraphics[width=0.72\textwidth]{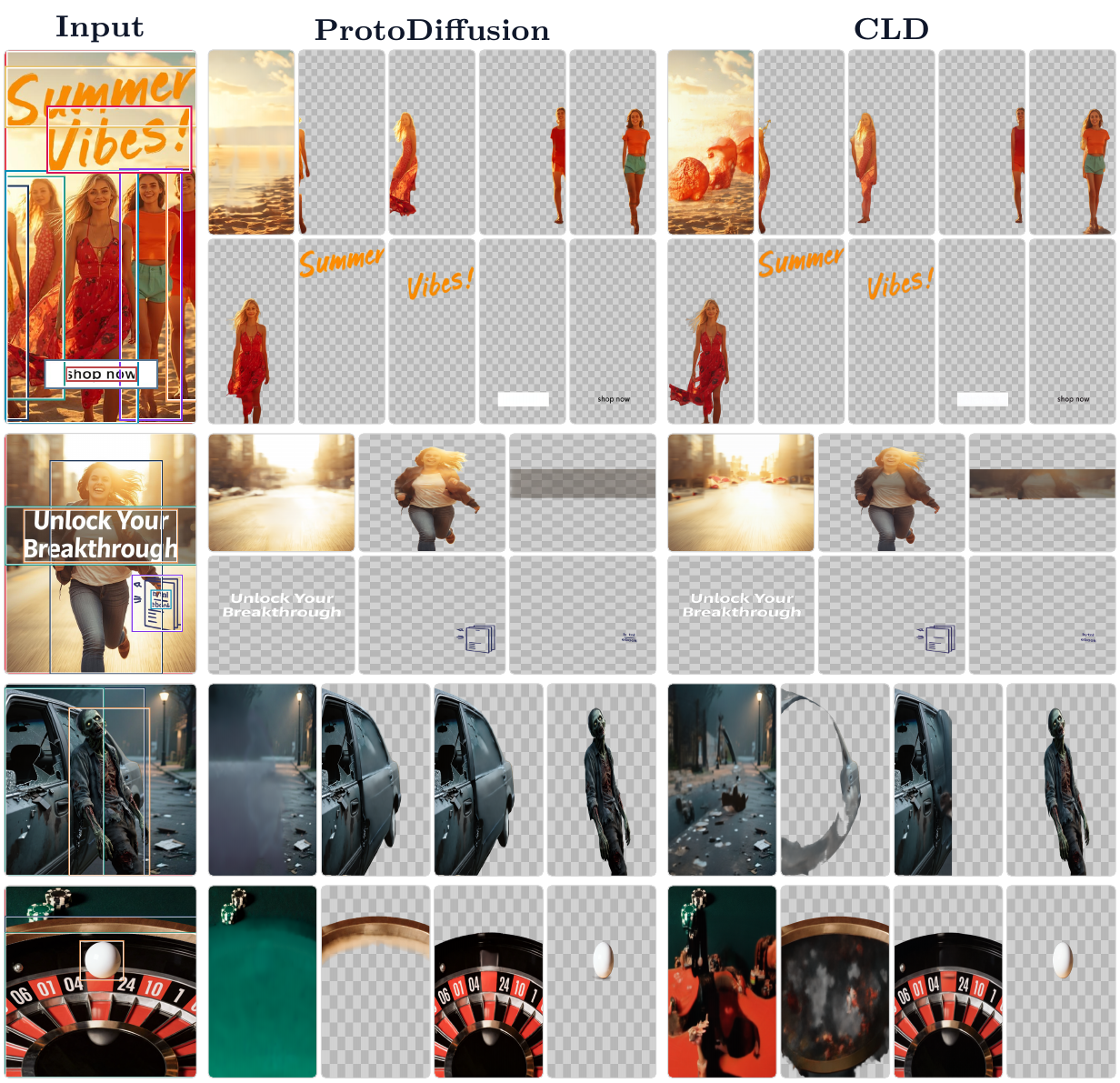}
\caption{Qualitative box-guided image-to-design decomposition on Crello
inputs. Columns: the input composite with the supplied element boxes
overlaid, and the background and layers recovered by \paper{} and CLD,
shown over a checkerboard. Examples are selected and illustrative, not a
systematic comparison.}
\label{fig:comp-bbox}
\end{figure*}

\subsection{Experimental Setup}
\label{sec:exp-implementation}

\paragraph{Training datasets.}
We train on layered graphic designs and PrismLayersPro\blue{~\cite{chen2025prismlayers}} compositions, augmented with synthetic
compositions built from COYO\blue{~\cite{byeon2022coyo}} images and graphic elements. We also use RevealLayer\blue{~\cite{revealllayer2026}} data and its
foreground elements, together with additional synthetic examples generated using FLUX and LanPaint\blue{~\cite{zheng2025lanpaint}}.
ProtoVAE is trained on this mixture with random layer permutations. ProtoDiffusion uses the same
data for image-to-design training with and without bounding boxes. For text-to-design, the final
training stage uses a curated subset that excludes fully synthetic compositions to preserve aesthetics.

\paragraph{Evaluation datasets.}
We evaluate reconstruction and image-to-design decomposition on Crello\blue{~\cite{yamaguchi2021canvasvae}}, following the LayerD
evaluation split for decomposition~\cite{suzuki2025layerd}.

\textbf{Training schedule.}
To adapt the pretrained image backbone, we first train the newly introduced candidate,
cross-attention, and box-conditioning modules, then progressively unfreeze the backbone.
The supplementary material summarizes the training setup.

\textbf{Baselines.}
We compare image-to-design decomposition with
QIL~\cite{qwenimagelayered2025}, LayerD~\cite{suzuki2025layerd},
and CLD~\cite{liu2025cld}, and computational efficiency with
QIL and MRT~\cite{tang2026mrt}.

\subsection{Reconstruction Fidelity}
\label{sec:exp-recon}

\begin{table*}[tbp]
\centering
\small
\setlength{\tabcolsep}{4pt}
\caption{RGBA reconstruction on 1,000 held-out Crello designs (10,036 layers) at 1024 pixels on the longest axis.
Joint indicates shared compression of the layer stack. PSNR, SSIM\blue{~\cite{wang2004ssim}}, and LPIPS\blue{~\cite{zhang2018lpips}} average nine backgrounds, then layers and designs; rFID\blue{~\cite{heusel2017fid}} averages background-specific distribution scores.
QIL + adapted decoder is an independent-layer reference.
Best values overall are bold; best joint-model values are underlined.
The bold row denotes our best joint reconstruction model.}
\label{tab:planned-vae}
\resizebox{\textwidth}{!}{%
\begin{tabular}{@{}lcrrrrrrr@{}}
\toprule
& & \multicolumn{3}{c}{Full layer canvas} & \multicolumn{3}{c}{GT crop} & \\
\cmidrule(lr){3-5}\cmidrule(lr){6-8}
Autoencoder / variant & Layered & PSNR$\uparrow$ & SSIM$\uparrow$ & LPIPS$\downarrow$ & PSNR$\uparrow$ & SSIM$\uparrow$ & LPIPS$\downarrow$ & rFID$\downarrow$ \\
\midrule
LayerDiffuse RGBA & $\times$ & 42.04 & 0.9749 & 0.01706 & 28.97 & 0.8701 & 0.04129 & 2.432 \\
AlphaVAE (FLUX) & $\times$ & 35.57 & 0.9736 & 0.02773 & 27.70 & 0.8833 & 0.05287 & 30.669 \\
QIL RGBA-VAE & $\times$ & 49.59 & 0.9911 & \textbf{0.00324} & 37.83 & 0.9744 & \textbf{0.00659} & 1.477 \\
\midrule
QIL + adapted decoder & $\times$ & \textbf{54.94} & \textbf{0.9956} & 0.00391 & \textbf{41.92} & \textbf{0.9882} & 0.00683 & \textbf{0.326} \\
ProtoVAE AdaLN-Zero decoder & $\checkmark$ & 51.90 & 0.9943 & 0.00499 & 39.17 & 0.9842 & 0.00862 & 1.042 \\
ProtoVAE Mixed AdaLN & $\checkmark$ & 51.89 & 0.9944 & 0.00480 & 39.04 & 0.9856 & \underline{0.00787} & 1.251 \\
\textbf{ProtoVAE AdaLN-Zero + GAN + leakage} & $\checkmark$ & \underline{53.31} & \underline{0.9951} & \underline{0.00448} & \underline{40.43} & \underline{0.9863} & 0.00806 & \underline{0.786} \\
ProtoVAE + ConvFFN4 & $\checkmark$ & 50.54 & 0.9940 & 0.00657 & 38.27 & 0.9775 & 0.01248 & 1.220 \\
\bottomrule
\end{tabular}}
\end{table*}

ProtoVAE achieves high-fidelity joint reconstruction (Table~\ref{tab:planned-vae}).
Our best joint model reaches 40.43\,dB crop PSNR and an rFID of 0.786, the lowest among
joint models. The independent-layer reference reaches 41.92\,dB crop PSNR.
The supplementary material provides model definitions and evaluation details.

\subsection{Design Decomposition}
\label{sec:exp-genquality}

\begin{table*}[tbp]
\centering
\small
\setlength{\tabcolsep}{4.5pt}
\caption{Image-to-design decomposition on the Crello test split.
We use the LayerD/QIL metrics with our composition-preserving alignment
for all methods, including detector-based baselines.
RGB L1 ($\downarrow$) measures color error weighted by ground-truth alpha;
alpha soft IoU ($\uparrow$) measures overlap between predicted and
ground-truth alpha maps. Columns specify the maximum number of allowed
adjacent-layer merges. Our model is evaluated with and without supplied
bounding boxes. QIL is fine-tuned on the Crello training split.
The best and second-best results in each column are bold and underlined,
respectively; ties receive the same marking.}
\label{tab:planned-diffusion}
\begin{tabular}{@{}lcccccccccccc@{}}
\toprule
& \multicolumn{6}{c}{RGB L1$\downarrow$} & \multicolumn{6}{c}{Alpha soft IoU$\uparrow$} \\
\cmidrule(lr){2-7}\cmidrule(lr){8-13}
\# max allowed layer merges & 0 & 1 & 2 & 3 & 4 & 5 & 0 & 1 & 2 & 3 & 4 & 5 \\
\midrule
VLM Base $+$ Hi-SAM & 0.120 & 0.103 & 0.089 & 0.081 & 0.076 & 0.073 & 0.560 & 0.630 & 0.686 & 0.722 & 0.747 & 0.759 \\
YOLO Base $+$ Hi-SAM & 0.096 & 0.083 & 0.071 & 0.063 & 0.059 & 0.058 & 0.570 & 0.654 & 0.717 & 0.757 & 0.781 & 0.790 \\
LayerD & 0.071 & 0.054 & 0.046 & 0.042 & 0.040 & 0.040 & 0.752 & 0.811 & 0.844 & 0.856 & 0.862 & 0.865 \\
QIL & 0.059 & \secbest{0.049} & \secbest{0.039} & \secbest{0.038} & \secbest{0.036} & \secbest{0.036} & \best{0.871} & \best{0.886} & \best{0.911} & \secbest{0.912} & 0.916 & 0.916 \\
CLD & 0.081 & 0.075 & 0.063 & 0.060 & 0.055 & 0.053 & 0.822 & 0.857 & \secbest{0.897} & \best{0.913} & \best{0.929} & \best{0.935} \\
\midrule
\paper & \secbest{0.057} & \secbest{0.049} & 0.045 & 0.042 & 0.040 & 0.037 & 0.681 & 0.760 & 0.821 & 0.860 & 0.884 & 0.904 \\
\paper{} (with bbox) & \best{0.040} & \best{0.034} & \best{0.030} & \best{0.029} & \best{0.028} & \best{0.027} & \secbest{0.850} & \secbest{0.871} & \secbest{0.897} & 0.909 & \secbest{0.922} & \secbest{0.930} \\
\bottomrule
\end{tabular}
\end{table*}

Table~\ref{tab:planned-diffusion} evaluates layer recovery using the LayerD/QIL metrics:
RGB L1 weighted by ground-truth alpha and alpha soft IoU, with zero to five adjacent-layer
merges. Figures~\ref{fig:comp-nobbox} and~\ref{fig:comp-bbox} show corresponding qualitative results.

We replace the monotonic DTW matching used in LayerD/QIL with
composition-preserving alignment: we search render-preserving layer groupings
and match them using a minimum-mean-cost bipartite edge cover. This accommodates
equivalent orders of non-overlapping layers and the finer or coarser
decompositions produced when our model operates without a prescribed layer
count or bounding boxes (see the supplementary material).
Per-layer metrics distinguish editable decomposition from simply reproducing the input
in a single opaque layer.

With boxes, \paper achieves the lowest RGB L1 at every merge allowance, reaching 0.040
without merges and 0.027 with five merges. Without boxes, its zero-merge RGB L1 of 0.057
is slightly lower than QIL's, while alpha soft IoU remains lower throughout the merge sweep.
QIL leads alpha soft IoU at zero to two merges and CLD at three to five, distinguishing
foreground color fidelity from agreement with ground-truth layer support.

\subsection{Text-to-Design Generation}
\label{sec:exp-t2d-qual}

For merged-image quality, we evaluate 1,000 text-only prompts: 400 general
design briefs, 100 each featuring architecture, food, humans, animals, and
nature, and 100 covering art, products, and miscellaneous subjects. We use
four seeds per prompt for each method (4,000 attempted generations per method).
For prompt adherence, text accuracy, and runtime, we use 50 newly authored
briefs with one generation per prompt.
\paper{} generates editable RGBA layers, which we merge into an RGB image for
comparison with the RGB-only FLUX.2 [klein]~\cite{flux2klein2025} and
Qwen-Image~\cite{qwenimage2025} baselines. FLUX.2 [klein] also serves as our
pretrained RGB backbone.
We evaluate merged RGB outputs using FID and paired-reference
SSIM and PSNR. Prompt adherence and prescribed-text accuracy
are model-assessed. Runtime is the median batch-one end-to-end
time on H200 GPUs, including cascade decomposition and excluding
model loading and evaluation. Reference data, scoring procedures,
and inference settings are detailed in the supplementary material.

\begin{table}[tbp]
\centering
\small
\setlength{\tabcolsep}{2.5pt}
\caption{Text-to-design results. Merged RGB quality uses the design prompts;
prompt adherence, text accuracy, and runtime use the briefs. FID is
distributional; SSIM uses paired references. Align. and OCR are model-assessed
scores in $[0,1]$. Layered indicates whether the method produces separate
RGBA layers. Time is median batch-one end-to-end runtime on H200 GPUs,
excluding model loading and evaluation. FLUX denotes FLUX.2 [klein]; QI
denotes Qwen-Image. QIL denotes the decomposer. Dashes indicate unmeasured
values. Best and second-best values are bold and underlined, respectively.}
\label{tab:planned-t2d}
\resizebox{\columnwidth}{!}{%
\begin{tabular}{@{}lcccccc@{}}
\toprule
& & \multicolumn{2}{c}{Merged RGB} & \multicolumn{3}{c}{Prompt and runtime} \\
\cmidrule(lr){3-4}\cmidrule(lr){5-7}
Method & Layered & SSIM$\uparrow$ & FID$\downarrow$ & Align.$\uparrow$ & OCR$\uparrow$ & Time (s)$\downarrow$ \\
\midrule
QI & $\times$ & 0.5904 & 98.08 & \textbf{0.984} & \underline{0.850} & 19.31 \\
QI + QIL & $\checkmark$ & --- & --- & \textbf{0.984} & 0.840 & 230.80 \\
FLUX & $\times$ & \secbest{0.5925} & \secbest{86.84} & 0.848 & 0.460 & \textbf{9.63} \\
FLUX + QIL & $\checkmark$ & --- & --- & 0.832 & 0.470 & 220.86 \\
\midrule
\paper{} & $\checkmark$ & \best{0.6898} & \best{66.41} & \underline{0.940} & \textbf{0.860} & \underline{11.58} \\
\bottomrule
\end{tabular}}
\end{table}

Table~\ref{tab:planned-t2d} shows that \paper{} has the highest merged-image
SSIM and the lowest FID, ahead of the RGB-only FLUX.2 [klein] and Qwen-Image baselines.
Because these baselines produce no layers, we also decompose their outputs with
QIL to compare complete layer-producing pipelines. \paper{} reaches 0.940 alignment
and the highest OCR score (0.860); Qwen-Image and its QIL cascade have
higher alignment. \paper{}'s 11.58\,s median runtime is $19.1\times$ shorter than
the FLUX.2 [klein] + QIL cascade and $19.9\times$ shorter than the
Qwen-Image + QIL cascade.

\subsection{Computational Efficiency}
\label{sec:exp-efficiency}
\label{sec:exp-ablations}

\begin{figure}[!ht]
  \centering
  \includegraphics[width=\columnwidth]{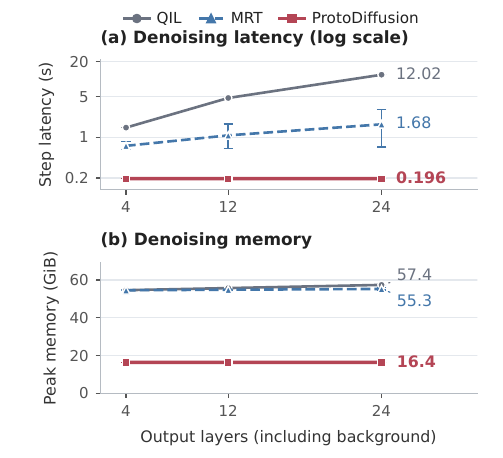}
  \caption{Layer-count scaling at $640\times640$ on an H200.
  Points show mean denoising-step latency and peak allocated memory across three layouts;
  whiskers show their range. Counts include the background; \paper{} uses $M=N$.
  Native systems differ in backbone and conditioning; output quality is evaluated separately.}
  \label{fig:measured-layer-scaling}
\end{figure}

Figure~\ref{fig:measured-layer-scaling} reports denoising cost for
4, 12, and 24 output layers. From 4 to 24 layers, mean step latency
increases from 1.48 to 12.02 seconds for QIL and from 0.72 to
1.68 seconds for MRT, while \paper{} remains near 0.20 seconds.
MRT and \paper{} receive boxes, whereas QIL receives the layer count.
These measurements compare native systems with different backbones
and conditioning; they are not quality-matched or controlled
backbone comparisons. Further protocol details are provided in
the supplementary material.

\section{Limitations}
\label{sec:discussion}

Capacity for very dense or heavily overlapping compositions remains to be characterized.
Preserving thin elements and reducing residual leakage remain challenges, while RGBA
decoding cost still grows with the number of output layers.

Despite strong RGB fidelity, box-free alpha soft IoU remains below that of QIL.
Controlled ablations would isolate individual component contributions.

Broader natural-image evaluation and occupancy calibration across domains would further
characterize the framework's generality.

\section{Conclusion}
\label{sec:conclusion}

\paper introduces a token-per-layer representation that combines a shared spatial prototype
with one candidate vector per RGBA layer. Joint denoising supports image-to-design with or
without bounding boxes and text-to-design within a common architecture. Experiments demonstrate
faithful reconstruction, accurate layer recovery, and favorable merged-image
quality, text accuracy, and runtime for text-to-design generation.
This factorization offers a compact basis for generating and decomposing editable
layered designs.

\label{cvpr-main-body-end}
{
    \small
    \bibliographystyle{ieeenat_fullname}
    \bibliography{main}
}
\clearpage
\appendix
\section*{Supplementary Material}

\section{Representation and Training Details}
\subsection{\protovae: token-per-layer latents}
\label{sec:supp-method}

Given a design with $N$ RGBA layers $\{x_i\}_{i=1}^{N}$ (including the full-canvas background)
and their normalized bounding boxes $\{b_i\}_{i=1}^N$, \protovae encodes:

\begin{itemize}
  \item a \textbf{prototype} $p \in \mathbb{R}^{C \times H \times W}$: a single spatial latent grid
  (e.g.\ $H{=}W{=}32$) that encodes the shared spatial context of the layered design, produced once
  per design, independent of $N$;
  \item $N$ \textbf{candidate vectors} $c_i \in \mathbb{R}^{C}$: one contextualized vector per layer,
  conditioning its reconstruction from the shared prototype.
\end{itemize}

For \protovae encoding, each $b_i$ is determined directly from layer $i$'s own alpha support and
therefore requires no external layout annotation. This differs from \paper's optional
bounding-box conditioning (Section~\ref{sec:method-diffusion}), which provides generation-time
optional spatial guidance during generation.

The encoder and reconstruction path are shown in Figure~\ref{fig:protovae-architecture}.

\paragraph{Token-per-layer factorization.} The frozen QIL encoder maps each input layer to a spatial
latent, which is projected to hidden width $D$ to obtain $z_i\in\mathbb{R}^{D\times H\times W}$.
A learned token $z_\ell\in\mathbb{R}^{D}$ is broadcast over the spatial grid and appended as an
additional layer. LayerSquasher jointly processes these $N{+}1$ maps and preserves their spatial
resolution. The layer-token output parameterizes the prototype posterior. Each remaining output
is cropped to a region around its layer's alpha-derived bounding box, resized, and summarized by
a small CNN with global average pooling. An attention-based distribution extractor refines the
pooled features using prototype context and box features to parameterize the candidate posterior.

Both are parameterized as diagonal-Gaussian posteriors with $C{=}128$ channels.
During ProtoVAE training, reparameterized samples are used for reconstruction, and KL
terms regularize both posteriors toward a standard normal prior. Diffusion training uses
the exported posterior means (equivalently, modes), rather than resampling the posteriors.
A decoder reconstructs each layer's
QIL latent by conditioning the shared prototype decoder on candidate $c_i$ through AdaLN-Zero; the reconstructed per-layer latents are then passed through the QIL RGBA
decoder~\cite{qwenimagelayered2025} to obtain pixels. The QIL decoder is trained jointly with ProtoVAE,
while the QIL encoder remains frozen. Crucially, \textbf{the candidate tokens are the only part
of the representation whose count scales with $N$} --- the prototype grid is shared and paid for
once, independent of how many elements the design has.

\paragraph{Contextualized candidate vectors.} Each candidate starts from its own
post-LayerSquasher feature map, cropped to a region around that layer's alpha-derived bounding box, resized, and summarized
by a small CNN with global average pooling. An attention-based distribution extractor then combines
the pooled features with prototype context and box features, with training-time dropout, to
parameterize the posterior for $c_i$. Because LayerSquasher
jointly attends across all $N{+}1$ inputs, each candidate's CNN input has already incorporated the
surrounding composition and the other layers before this encoding/refinement stage.

\paragraph{AdaLN-Zero decoder.}
Each candidate modulates the shared spatial prototype through AdaLN-Zero~\cite{peebles2023dit},
providing a direct conditioning path throughout the decoder. Mixed-modulation variants
also concatenate candidate and prototype tokens (Appendix~\ref{sec:recon-pilot-protocol}).

\paragraph{Transparency representation.} \blue{RGB is not premultiplied by
alpha; transparent pixels receive a deterministic padded RGB value, and
reconstruction uses spatially weighted $L_1$ terms.}

\paragraph{Reconstruction objectives.}
Pixel and latent reconstruction terms emphasize layer isolation in addition to the final
composition. The reported configurations change several components together; individual
effects require controlled ablations.

\paragraph{Prototype alignment regularization.}
Inspired by DC-Gen's embedding alignment for transferring pretrained diffusion models across
latent spaces~\cite{he2025dcgen}, we regularize the prototype against the frozen QIL encoding
of the alpha-composited image. An alignment module maps both representations to comparable features,
and a mean-squared-error term encourages agreement between them. This anchors the shared spatial
component of ProtoVAE to the original QIL representation, with the aim of reducing the
representation gap and facilitating convergence when adapting the pretrained diffusion backbone.
Layer reconstruction supervises hidden content separately, with increased weight in occluded regions;
alignment does not require the prototype to equal the composite latent.

\paragraph{Overlap-weighted and inter-element relational losses.} Both pixel- and latent-space
losses are re-weighted toward two regions that can be under-represented in an average reconstruction loss: regions of high inter-layer overlap, and regions just outside an element's own
footprint where visibility leakage (a shadow/halo artifact) appears (alpha-only weighting there, since
RGB has no visible contribution in fully transparent regions). We additionally add an \emph{inter-element relational
loss}: per-layer latent features are pooled using alpha weights, and their pairwise cosine
similarities are compared with those of the corresponding ground-truth layers. An asymmetric
penalty targets excess similarity between reconstructed layers.

\paragraph{Adversarial losses.} An attention-based latent-space discriminator sees stacked per-layer
design latents; a per-element PatchGAN~\cite{isola2017pix2pix} sees local alpha-composited crops (predicted
vs.\ ground-truth element over a shared random background, resized so small elements get high-resolution
patches and large/many-element designs get zoomed-out ones), filtered toward element boundaries and
high-structural-difference regions. Both use spectral normalization~\cite{miyato2018spectralnorm} and
a staged warm-up schedule for stability.

\subsection{The \paper transformer}
\label{sec:supp-diffusion}

The \paper transformer adapts FLUX.2-Klein-4B~\cite{flux2klein2025} to jointly
 denoise a shared prototype grid and layer-specific candidate tokens. Dedicated interacting
streams combine spatial context and layer identity under image or text conditioning.
The architecture extends the pretrained backbone in three ways:

\paragraph{Optional bounding-box-aware RoPE.}
Prototype tokens use spatial positional encoding. When boxes are available, candidate
positional encoding and Fourier features convey their location and extent; without boxes,
candidate IDs identify the slots.

\paragraph{Dedicated candidate stream.}
Prototype tokens use the pretrained image stream, while candidates have a dedicated
self-attention stream. Bidirectional cross-attention periodically exchanges information
between spatial context and layer identities.

\paragraph{Conditioning settings.}
Image-to-design uses the encoded composite as conditioning.
Text-to-design jointly samples the prototype and candidates from noise under text conditioning. Optional boxes provide spatial guidance
in either setting. All modes use a fixed candidate capacity, with an activity threshold selecting
the retained layers as described in Section~\ref{sec:method-diffusion}.

\paragraph{Slot order.}
Layers are assigned to slots by increasing $z$-layer (back to front). Within each level,
box centers are sorted lexicographically by $(y,x)$, giving top-to-bottom, then left-to-right
order. Candidate targets and box metadata follow the same permutation. Empty slots are appended;
filtering at inference preserves the relative order of retained slots for compositing.
\paragraph{Training objective: flow matching and QIL-space layered losses.}
Flow matching~\cite{liu2022rectifiedflow,lipman2023flowmatching} uses
$z_t=(1-t)z_0+t\epsilon$ and predicts $v=\epsilon-z_0$, with $z_0$ in diffusion
coordinates. For one design, let $e_p$ and $e_c$ denote predicted-minus-target velocity errors.
The separately reduced stream losses are
\[
\begin{aligned}
\mathcal{L}_p&=\frac{1}{CHW}\sum_k w^p_k e_{p,k}^{\,2},\\
\mathcal{L}_c&=\frac{1}{NC}\sum_{i\le N,d}e_{c,id}^{\,2}
+\frac{\eta}{(M-N)C}\sum_{i>N,d}e_{c,id}^{\,2},
\end{aligned}
\]
where an empty summation group contributes zero and $\eta=1.5$.
The batch loss averages per-design losses. Thus increasing the number of prototype entries
or padding slots does not implicitly increase their contribution. The candidate coefficient is
$\lambda_c$, independently of any layered-loss coefficients.

\paragraph{Occlusion weighting.}
As in ProtoVAE, spatial weights emphasize information hidden by other layers.
The prototype-flow weighting path uses $w^p_k=1+\gamma o_k$ in image-to-design, where $o_k$
counts overlapping alpha supports beyond the first, downsampled to the prototype grid;
otherwise $w^p_k=1$.
The layered reconstruction term uses $w_{ik}=1+\beta m_{ik}$, where $m_{ik}\in[0,1]$ is
layer $i$'s occlusion mask downsampled to its QIL grid and broadcast over channels.
Weights multiply errors before averaging and are not normalized away.
The full-layer target includes hidden content, whereas prototype alignment in ProtoVAE
regularizes shared features toward the visible composite. These are distinct supervision signals.

\paragraph{Layer reconstruction and separation.}
From the predicted velocity we recover $\hat z_0=z_t-t\hat v$ and decode
$(\hat p,\hat c_i)$ through frozen ProtoVAE to full-canvas QIL latents $\hat u_i$.
The selected reconstruction objective is weighted $L_1$:
\[
\mathcal{L}_{\mathrm{QIL\text{-}L1}}=\frac{1}{NK}
\sum_{i=1}^{N}\sum_{k=1}^{K}w_{ik}|\hat u_{ik}-u_{ik}|.
\]
Padding slots are excluded. The contrastive formulation is
\[
\mathcal{L}_{\mathrm{sep}}=-\frac{1}{N}\sum_{i=1}^{N}
\log\frac{\exp(s(\hat u_i,u_i)/\tau)}
{\sum_{j=1}^{N}\exp(s(\hat u_i,u_j)/\tau)}.
\]
Here $s$ is cosine similarity between flattened full-canvas QIL latents: flattening preserves
coordinate correspondence, unlike spatial pooling or recentering element crops.
Consequently, identical-looking elements at different locations have different targets.
This formulation is inspired by RevealLayer's separation objective~\cite{revealllayer2026};
it does not reproduce its RGB-space orthogonality loss.

\paragraph{Weights and training phases.}
The total diffusion loss is
$\mathcal{L}_p+\lambda_c\mathcal{L}_c+
\lambda_{\mathrm{layer}}(\mathcal{L}_{\mathrm{QIL\text{-}L1}}+
\lambda_{\mathrm{sep}}\mathcal{L}_{\mathrm{sep}})$.
ProtoVAE is trained first and then frozen. Its alignment and adversarial objectives are
autoencoder-training terms, not additional diffusion losses. Diffusion training first adapts
the new modules, then progressively unfreezes the backbone as described below.

\subsection{Implementation details}
\label{sec:supp-implementation}
\label{sec:method-training}

The \paper transformer adapts FLUX.2-Klein-4B~\cite{flux2klein2025} with progressive
unfreezing: newly introduced modules train first, followed by backbone adaptation.
We use BF16 training with AdamW. The default candidate capacity is $M=50$;
image and text conditioning are dropped during training for classifier-free
guidance~\cite{ho2022cfg}. Separate loss weights balance prototype, candidate,
and layered supervision.

\subsection{Denoiser complexity and token-count study}
\label{sec:supp-token-analysis}

\paragraph{Two-stream denoiser cost.}
Let $S$ denote the number of prototype tokens after spatial packing, with one token per candidate.
For equal packing of full layer grids, their combined spatial sequence would contain $NS$ tokens.
Let $L_p,L_c$ count self-attention blocks in the prototype and candidate streams, and let $J$
count bidirectional exchange sites (one every $k=4$ blocks). Excluding conditioning tokens,
the number of query--key pairs is
\[
A=L_pS^2+L_cM^2+2JSM.
\]
The terms correspond to prototype self-attention, candidate self-attention, and both directions
of cross-attention. For a matched common hidden width $d$, the leading arithmetic cost is
\[
O\!\left(dA+d^2[L_pS+L_cM+J(S+M)]\right),
\]
including token-wise attention projections and MLPs. A dense full-grid transformer of depth $L$
instead costs $O(L[(NS)^2d+NSd^2])$. A region-grid model replaces $NS$ with the sum of its
packed region sizes; cropping can substantially reduce this quantity. Fixed-length conditioning
adds its attention interactions without changing the per-layer representation increment.

These expressions hold depth, widths, packing, and conditioning fixed to isolate the representation.
For growing $M\approx N$, candidate attention remains quadratic in capacity; the benefit is replacing
an additional spatial grid by one token, not eliminating all layer-dependent work or changing every
term's asymptotic class. At fixed $M$, changing the retained count $N\le M$ leaves denoising
sequence lengths unchanged. Actual FLOPs depend on each backbone's widths and blocks, and latency
also depends on kernels and guidance execution. Thus the analysis establishes structural scaling;
the measured system-level speedup is not a causal estimate of the factorization alone.

\begin{figure*}[tbp]
  \centering
  \includegraphics[width=0.86\linewidth]{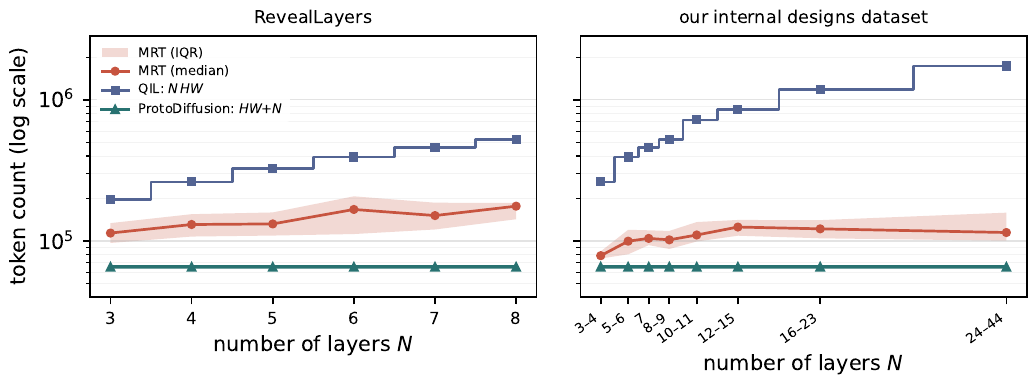}
  \caption{Representation token counts before packing at $2048\times2048$ on RevealLayers
  (left) and internal designs (right). ProtoVAE adds one candidate token per layer;
  QIL adds a full spatial grid. MRT scales with region area, shown as the median and
  interquartile range at each layer count. Conditioning and layout estimation are excluded.}
  \label{fig:token-scaling}
\end{figure*}

\paragraph{Token-count comparison.}
We compare representation token counts before backbone-specific packing using $HW+N$ for
ProtoVAE, $NHW$ for full layer grids, and $\sum_i\lceil a_i HW\rceil$ for region crops
of normalized area $a_i$. Conditioning and layout estimation are excluded; diffusion uses
$M$ candidate slots in place of the $N$ encoded layers.
Figure~\ref{fig:token-scaling} compares overlap-rich RevealLayers layouts with an internal
corpus of graphic designs at $2048\times2048$. Each corpus contains 250 layouts; the internal
corpus is non-public. These distinct layout distributions characterize representation scaling,
while runtime is measured separately. ProtoVAE adds one token per layer, full-grid methods add
$HW$, and region-based counts depend on element area.
The internal corpus contains more, smaller elements: the median token-count ratio relative to
ProtoVAE increases from $4.0\times$ to $9.0\times$ for QIL and decreases from $1.96\times$ to
$1.64\times$ for MRT. Figures~\ref{fig:supp-token-distribution-2048}
and~\ref{fig:supp-token-distribution-art-2048} show the corresponding distributions.

\section{Additional Evaluation and Context}
\subsection{Candidate-score calibration}
\label{sec:candidate-calibration}

We calibrate candidate activity on 1,954 examples, split equally between calibration and
evaluation. The score is
$a_i=|\operatorname{mean}(\tilde c_i)|+\operatorname{std}(\tilde c_i)$, computed across channels of the denoised candidate vector $\tilde c_i$
before filtering. We select $\theta$ by minimizing layer-count MAE against ground truth on the
calibration split, choosing the lowest threshold in a tie. Candidates with $a_i>\theta$ are
retained, with a minimum of one candidate. Ground-truth counts include the background.

\begin{figure}[tbp]
  \centering
  \includegraphics[width=\linewidth]{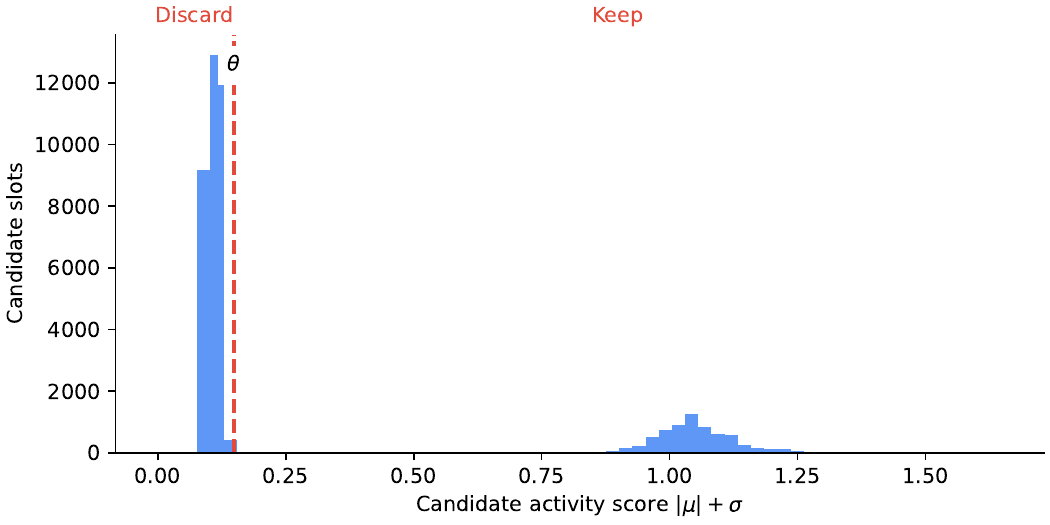}
  \caption{Candidate activity scores on the evaluation split. The dashed line marks the
  threshold $\theta=0.1494$, selected on the calibration split. Scores below the cutoff are
  discarded, except that the highest-scoring candidate is always retained.}
  \label{fig:crello-candidate-scores}
\end{figure}

The selected threshold yields an evaluation count MAE of 3.223
(95\% bootstrap interval $[3.014,3.439]$) and exact-count accuracy of 12.8\%, compared with
MAE 3.275 at the original cutoff of $0.75$. Although the score distribution is bimodal,
77.99\% of evaluation samples retain fewer layers than the ground truth. Threshold calibration
therefore does not resolve differences in decomposition granularity or establish semantic layer
coverage. These post-hoc results require independent validation.

\subsection{Measured layer-count scaling}
\label{sec:measured-latency-protocol}
We evaluate nine procedural layouts: 4, 12, or 24 output layers (including
background), each with foreground bounding-box areas of 1\%, 8\%, or 25\%.
This benchmark sets ProtoDiffusion's candidate capacity to the requested count ($M=N$),
rather than measuring the default fixed $M=50$ configuration.
All systems use the same $640\times640$ inputs, seed 42, 20 steps, guidance 1.5,
batch size one, and BF16 diffusion weights. Runs execute sequentially on an
exclusively reserved H200, without quantization, compilation,
offloading, or denoiser activation caching. Each iteration is CUDA-synchronized;
each layout is run once, and we discard the first five steps and average the remaining fifteen. Whiskers
represent variation across layouts, not confidence intervals.

QIL uses the public Qwen-Image-Layered model. QIL and MRT use their native Qwen backbones, while ProtoDiffusion uses FLUX.2-Klein-4B. MRT and ProtoDiffusion receive bounding boxes; QIL receives the output count. QIL and
MRT share a fixed text prompt, while ProtoDiffusion's native image-to-design path omits
text tokens. Native schedulers and VAEs are retained. All runs use PyTorch 2.9.1;
MRT uses Diffusers 0.35.1, whereas QIL and ProtoDiffusion use 0.41.0.dev0. Thus the figure
measures deployed systems rather than an isolated representation ablation.
Native guidance execution differs: QIL and MRT use two sequential denoiser calls
per step, whereas ProtoDiffusion batches the conditional and unconditional branches.
Attention implementations are retained per system; the executed kernels were not profiled.

Memory denotes peak CUDA allocated memory during denoising, including model
weights, rather than reserved or total device memory. The accompanying CSV also
records the post-denoising peak and total generation-call latency, excluding model
loading and output-file writes. Layout/count estimation is excluded because these
inputs are supplied. Every run returns the requested number of $640\times640$
layers; this dimensional check does not establish semantic completeness or quality.
Raw measurements and the plotting script accompany the paper under
\texttt{figures/matched\_latency/} and \texttt{code/matched\_latency/}.

\subsection{Qualitative text-to-design examples}
\label{sec:supp-t2d-analysis}

\begin{figure*}[tbp]
  \centering
  \includegraphics[width=\textwidth]{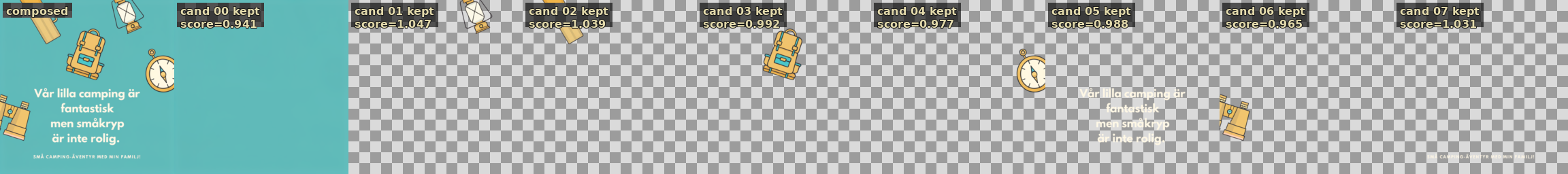}\\[-2pt]
  \includegraphics[width=\textwidth]{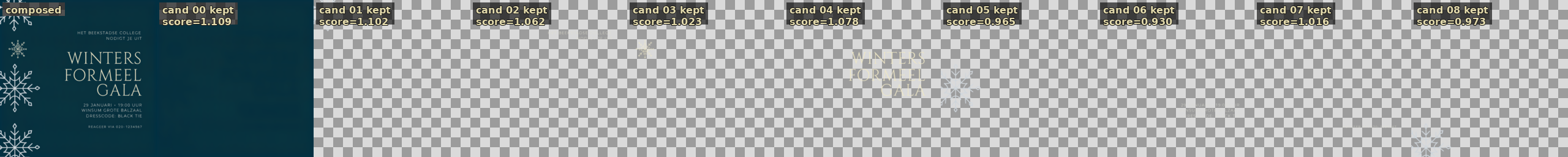}\\[-2pt]
  \includegraphics[width=\textwidth]{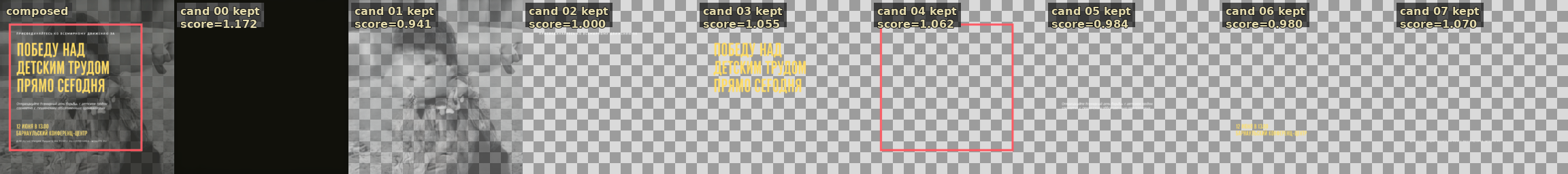}
  \caption{Selected text-to-design compositions and decoded RGBA layers. Illustration and
  typography examples show distinct editable elements; the photographic example illustrates
  remaining redundancy in background and text layers. Checkerboards show transparency;
  displayed scores are inspector annotations.}
  \label{fig:text2design-qual}
\end{figure*}

Text-to-design jointly generates the prototype and candidates, whose decoded layers form the
composition. Figure~\ref{fig:text2design-qual} illustrates both visual coherence and the separation
of editable elements. Quantitative results appear in Section~\ref{sec:exp-t2d-qual}.

The selected examples illustrate clean separation of icons and decorative elements, alongside
challenges with low-opacity text and redundant layers in photographic compositions. Some outputs
also contain text-rendering errors. Empty slots can represent unused capacity or missed elements;
assessing this distinction requires element-level coverage evaluation.

\subsection{Extended related work}
\label{sec:supp-related}

\paragraph{Latent and transformer diffusion.} \blue{Diffusion models~\cite{ho2020ddpm}
generate images by iterative denoising; latent diffusion~\cite{rombach2022ldm}
moves the denoising into a compressed VAE latent space.} Diffusion Transformers
(DiT)~\cite{peebles2023dit} replace the U-Net backbone with a transformer and popularized AdaLN-Zero
conditioning, which we reuse both in \protovae's decoder and in the \paper transformer's candidate
stream (Sections~\ref{sec:method-protovae}--\ref{sec:method-diffusion}). \blue{Rectified flow~\cite{liu2022rectifiedflow} and flow matching with
linear (optimal-transport) paths~\cite{lipman2023flowmatching} replace the
diffusion SDE with straight noise-to-data interpolations trained by velocity
regression, the training objective of Stable Diffusion~3~\cite{esser2024sd3}
and other recent text-to-image transformers;}
\paper inherits this objective from its backbone and extends it with a per-layer consistency loss
(Section~\ref{sec:method-diffusion}). Modern open text-to-image
backbones such as Qwen-Image~\cite{qwenimage2025} and FLUX~\cite{flux2klein2025}
are the two families of pretrained transformers we
have adapted \paper from, via progressive unfreezing and RoPE~\cite{su2021roformer} extensions rather
than training from scratch; Appendix~\ref{sec:supp-diffusion} records the backbone choice.

\paragraph{Transparent / layered image diffusion.} LayerDiffuse~\cite{zhang2024layerdiffuse} \blue{extends} a pretrained latent diffusion backbone to emit a single transparent RGBA layer, by
adapting the VAE's latent space to jointly encode RGB and alpha (\emph{latent transparency}).
\qil~\cite{qwenimagelayered2025} is a recent model in this family: built
natively on Qwen-Image, its decomposition interface receives an image and a requested layer count.
Its decomposition interface returns the requested number of RGBA layers without supplied boxes.
The input image provides the composition. Its layer latents remain spatial grids, with attention
modeling interactions across layers.
\paper replaces the per-layer grid increment with a candidate vector.

\paragraph{Two-stage layout-then-content design generation, and MRT.} A separate line of work
generates graphic designs by first predicting an explicit layout and then rendering content into it.
\blue{COLE~\cite{jia2023cole} decomposes generation into hierarchical
subtasks handled by specialized large language/vision and diffusion models,
while Desigen~\cite{weng2024desigen} generates a background and then a layout
conditioned on it;} LayoutDM~\cite{inoue2023layoutdm} \blue{focuses on the layout-generation
sub-problem (bounding boxes and categories only), and LayoutGPT~\cite{li2023layoutgpt}
plans such layouts with an LLM before handing them to downstream generators.}
\textbf{MRT}~\cite{tang2026mrt} is a closely related system: a 20B-parameter masked-region diffusion model that unifies image-to-design (including natural images),
text-to-design, and layer addition/restyling --- tasks that motivate the layered setting. In our setup
(Section~\ref{sec:experiments}) MRT is evaluated as a two-stage pipeline for
transparent layered images: a \emph{region / layout-detection} model first proposes region boxes and
an element count, and a second \emph{layer-generate} diffusion model renders RGBA content into that
fixed layout, one call per full design. MRT thus conditions layered generation on a pre-specified
region layout, separating the prediction of \emph{how many elements, and where} from their visual
generation: an under- or over-segmented layout propagates unchanged into generation, and generating
$N$ layers pays a per-region spatial token cost that grows with the number and size of regions
(Section~\ref{sec:token-efficiency}). \paper instead offers candidate occupancy and optional box conditioning over a shared-grid representation.
The box-conditioned path uses external layout guidance; the no-box path does not require it.
\paragraph{Other layered-generation approaches.}
RevealLayer~\cite{revealllayer2026} targets natural-image decomposition and introduces
RevealLayerBench. PSDiffusion~\cite{huang2026psdiffusion} generates design layers concurrently
with global layer interactions. LayeringDiff~\cite{kang2025layeringdiff} generates a flat
image before decomposing it, while LayerD~\cite{suzuki2025layerd} uses iterative unoccluded-layer
extraction. SLEDGE~\cite{khan2026sledge} studies instruction-guided layered editing, and
\citet{wu2026syntheticlayered} investigate synthetic data for layered-design \blue{decomposition}.
These works address complementary aspects of generation, decomposition, and editing;
our focus is the shared-prototype representation and its joint diffusion model.

\paragraph{Design as structured latents / disentangled representations.} Our factorization of a
design into a shared context grid and per-element tokens is conceptually related to slot-based and
object-centric representation learning, \blue{and more generally to structured latents in the variational
autoencoder framework~\cite{kingma2014vae}.} Unlike object-discovery methods that must infer slots without any
supervision, \protovae has direct access to ground-truth per-layer RGBA at training time
(Section~\ref{sec:method-training}), from which each layer's spatial support is directly
observable, which we use to supervise an
\emph{inter-element relational loss} that supervises relationships between reconstructed
layer features (Section~\ref{sec:method-protovae}).

\paragraph{Adversarial and perceptual losses for reconstruction.} We use a PatchGAN-style
discriminator~\cite{isola2017pix2pix} with spectral normalization~\cite{miyato2018spectralnorm} on
alpha-composited local crops, and LPIPS~\cite{zhang2018lpips} for perceptual reconstruction loss, both
standard in image-to-image and autoencoder training but applied here to per-element alpha-composited
patches rather than whole images (Section~\ref{sec:method-protovae}).
\subsection{Reconstruction evaluation protocol}
\label{sec:recon-pilot-protocol}
We evaluate 1,000 randomly sampled Crello designs (10,036 layers), excluding the
100 designs used for model selection. Crello was not used to train our VAEs;
training-set overlap for external reference models is not independently verified.
All layers are retained. We resize the longest axis to 1024 pixels, preserve aspect
ratio, and transparently pad to multiples of 64; padding is removed before scoring.
Source images are resized to the evaluation resolution, including upsampling where needed.

Full-canvas and GT-alpha bounding-box crops (at least 32 pixels per axis, without
rescaling) use float32 predictions before image quantization. We compute PSNR,
Gaussian SSIM, and LPIPS-Alex over black, gray, white, red, yellow, green, cyan,
blue, and magenta backgrounds, averaging backgrounds, then layers within designs,
then designs. Empty-alpha layers use the full canvas. PSNR is capped at 120 dB.
rFID uses clean-fid Inception features with canonical uint8 preprocessing, computed
separately for each background over all 10,036 layers and then averaged.
All methods completed the evaluation. Additional metrics include
95\% confidence intervals from 5,000 paired design-bootstrap resamples, alpha L1,
soft alpha IoU, alpha-gradient L1, and GT-alpha-weighted RGB L1. Layers within a
design are not treated as independent bootstrap samples.

Inference is deterministic. LayerDiffuse uses FP32 and its native eight-way decoder
augmentation; other models use BF16. This reconstruction comparison is not a
matched-runtime benchmark. The reported variants are not controlled single-component ablations.

\paragraph{Variant definitions.}
ProtoVAE AdaLN-Zero decoder uses candidate-conditioned AdaLN-Zero modulation.
ProtoVAE Mixed AdaLN-Zero decoder combines AdaLN-Zero with concatenated candidate
and prototype tokens in the decoder.
GAN + leakage adds adversarial losses and leakage regularization.
ProtoVAE + ConvFFN4 decoder replaces decoder self-attention with convolutional feed-forward blocks.
QIL + adapted decoder retains independent per-layer encoding and uses the decoder
fine-tuned during ProtoVAE training, providing a reference without the shared
prototype--candidate bottleneck.

\section{Additional Qualitative Results}
\label{sec:supp-qual}

\subsection{Shared Prototype and Candidate-conditioned Layers}
\label{sec:supp-prototype-selection}

\begin{figure*}[tbp]
  \centering
  \includegraphics[width=\textwidth]{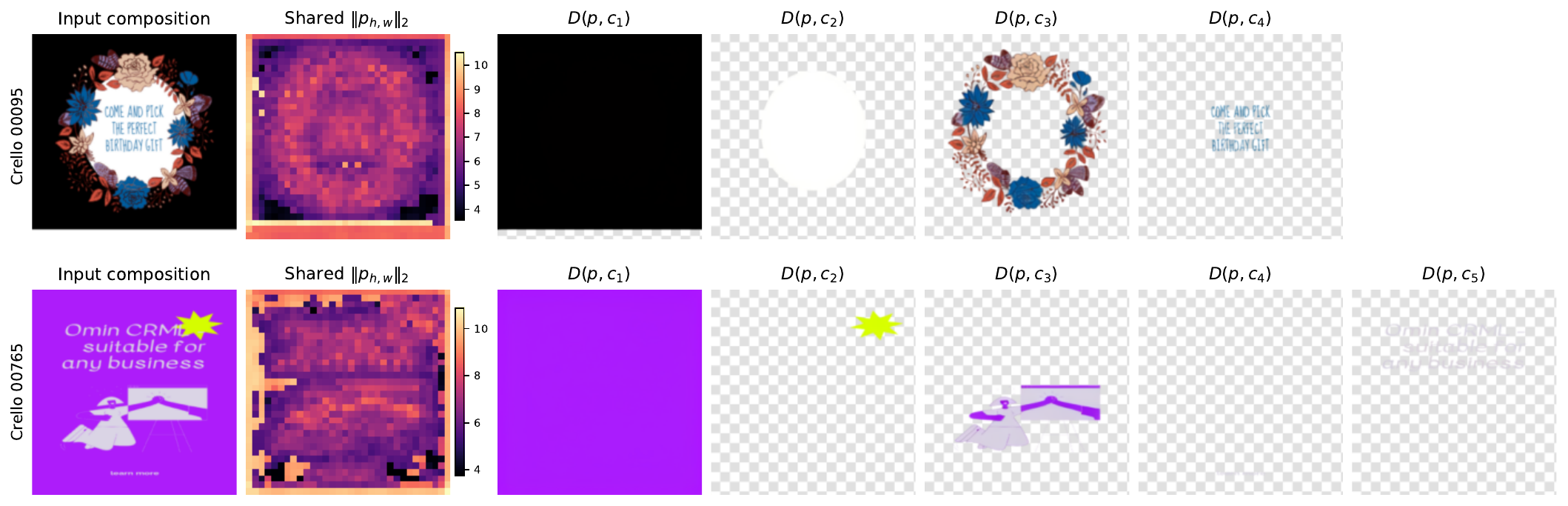}
  \caption{Shared prototype and layer selection on two designs.
  Left: input composition. Middle: $\|p_{h,w}\|_2$ across the 128 prototype channels
  (individual color scales). Right: all layers decoded with the same $p$ and each $c_i$;
  checkerboards indicate transparency. These are deterministic posterior-mode reconstructions
  with ProtoVAE, using inputs resized to $256\times256$ and a
  $32\times32$ prototype grid.}
  \label{fig:prototype-selection}
\end{figure*}

Figure~\ref{fig:prototype-selection} visualizes the channel-wise norm of a real encoded
prototype alongside layers reconstructed using that same prototype and their respective
candidates. Spatial structure is visible in the norm map, while candidate-conditioned
decoding recovers distinct RGBA elements. The norm is a feature-magnitude diagnostic;
it does not by itself establish how information is divided between the two components.

\subsection{Token-Count Distributions}
\label{sec:supp-token-distributions}

\begin{figure*}[tbp]
  \centering
  \includegraphics[width=0.76\textwidth]{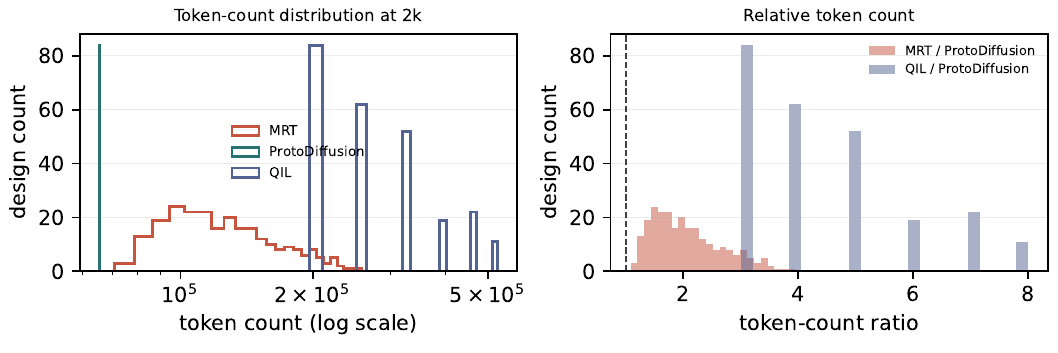}
  \caption{Per-design token-count distributions before packing at $2048{\times}2048$, using the same layouts and
  counting protocol as Figure~\ref{fig:token-scaling}.}
  \label{fig:supp-token-distribution-2048}
\end{figure*}

Figure~\ref{fig:supp-token-distribution-2048} complements the per-layer-count median/IQR summary in
Figure~\ref{fig:token-scaling} with the full per-design distribution, pooled across all $N$, at the
same $2048$px resolution. MRT's count varies continuously with the sum of crop areas, whereas QIL's
full-grid count changes discretely with the number of layers. Both are compared only with the
token count before packing of \paper; upstream layout estimation, when needed by MRT, is not included.

\subsection{Token-Count Distributions: an Internal Designs Corpus}
\label{sec:supp-token-art-dataset}

\begin{figure*}[tbp]
  \centering
  \includegraphics[width=0.76\textwidth]{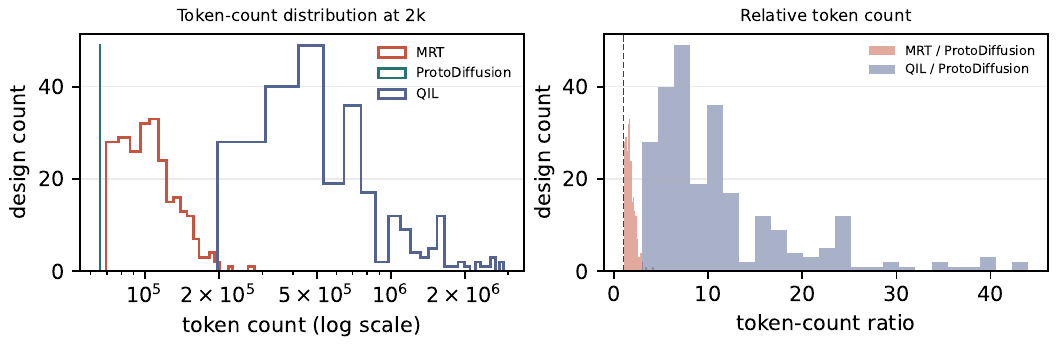}
  \caption{Per-design token-count distributions before packing at $2048{\times}2048$ for the same internal
  designs dataset sample as the right-hand panel of Figure~\ref{fig:token-scaling}, using the same
  counting protocol as Figure~\ref{fig:supp-token-distribution-2048}.}
  \label{fig:supp-token-distribution-art-2048}
\end{figure*}

Figure~\ref{fig:supp-token-distribution-art-2048} shows the internal-design distribution
corresponding to Figure~\ref{fig:token-scaling}, using the same resolution and counting protocol.

\subsection{Box-conditioned Image-to-design}
\label{sec:supp-i2d-qual}

Table~\ref{tab:supp-i2d-examples} illustrates successful element separation and remaining
challenges with text recovery, background completion, and redundant layers. Examples were
selected by a VLM across evaluation axes and serve as qualitative diagnostics.

\begin{table*}[tbp]
  \centering
  \small
  \caption{Selected box-conditioned image-to-design examples with VLM annotations.}
  \label{tab:supp-i2d-examples}
  \begin{tabular}{p{0.12\textwidth}p{0.19\textwidth}p{0.57\textwidth}}
    \toprule
    Sample & VLM axis / outcome & Observed decomposition behavior \\
    \midrule
    sample\_0005 & Text + non-text: 5/5 & Text (``The'', ``KOL'', ``Engagement'', ``workshop'') is
    isolated in individual layers with clean alpha; graphic foreground is removed from the white
    background without visible remnants. \\
    sample\_0007 & Non-text + inpainting: 5/5 & The forest scene remains in the background while
    the alligator, moon, trees, and roots are assigned to separate layers. The occluded background
    regions are described as semantically consistent, without holes or smears. \\
    sample\_0357 & Text: 0/5 & The large title (boxes 3--4) is absent from the foreground layers
    and survives only as a faint ghost in the background, with no dedicated text layers. \\
    sample\_0063 & Non-text: 0/5 & One decoded layer is a near-complete copy of the input, the background
    is a solid color rather than the scene, and the other decoded layers contain only small, misplaced
    fragments. \\
    sample\_0092 & Non-text + inpainting: 0/5 & Most decoded layers are empty; the remaining ones contain
    fragmented blue lines that do not match the supplied boxes. Since foreground extraction fails,
    the background is empty rather than inpainted. \\
    \bottomrule
  \end{tabular}
\end{table*}

\subsection{Text-to-design}
\label{sec:supp-t2d-qual}

\paragraph{Evaluation protocol.}
For the merged-image quality comparison, text prompts condition each generator.
The final RGB composites are evaluated with FID, SSIM, and PSNR. ProtoDiffusion
uses 20 denoising steps; the FLUX.2 [klein] and Qwen-Image RGB-only controls
use 50 steps each. \TODO{Document the prompt source and selection criteria,
real-image FID reference pool, paired-image SSIM/PSNR targets, valid sample
counts, seed aggregation, and metric preprocessing.} Training-data
disjointness of the briefs has not been independently verified; the reported
scores characterize this evaluation set.

The prompt-adherence, OCR, and runtime comparison uses newly authored briefs.
The RGB-only
FLUX.2 [klein] and Qwen-Image outputs are also decomposed with QIL to form
layer-producing cascades. Adherence and OCR scores are model-assessed. Runtime
is median batch-one end-to-end time on H200 GPUs, excluding model loading and
evaluation. Training-data disjointness of these briefs has not been
independently verified.

Additional selected text-to-design examples show the generated composites and decoded layers.
Checkerboards visualize transparency. Displayed scores and retained-slot labels are inspector
annotations rather than calibrated quality measures.

\begin{figure*}[tbp]
  \centering
  \begin{subfigure}[b]{0.31\textwidth}
    \includegraphics[width=\linewidth]{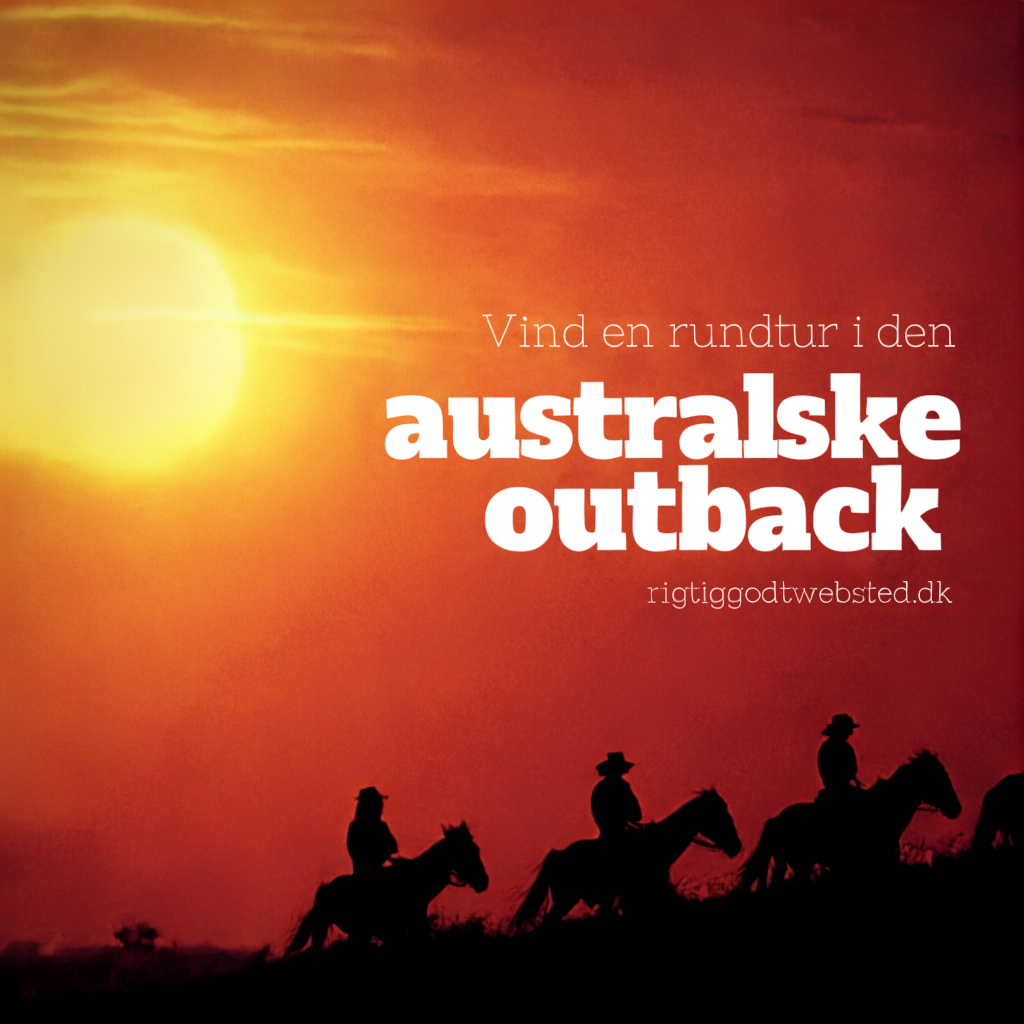}
    \caption{Outback travel poster.}
  \end{subfigure}\hfill
  \begin{subfigure}[b]{0.31\textwidth}
    \includegraphics[width=\linewidth]{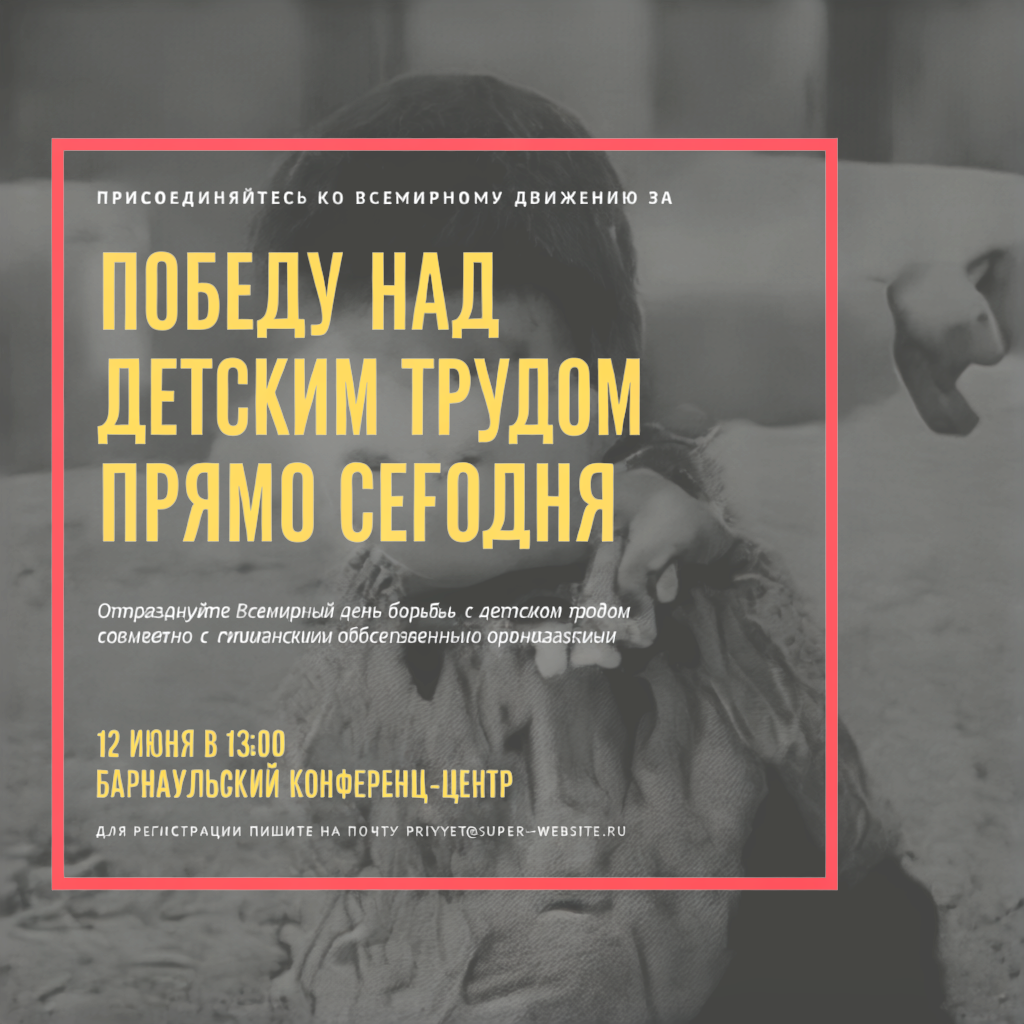}
    \caption{Awareness poster.}
  \end{subfigure}\hfill
  \begin{subfigure}[b]{0.31\textwidth}
    \includegraphics[width=\linewidth]{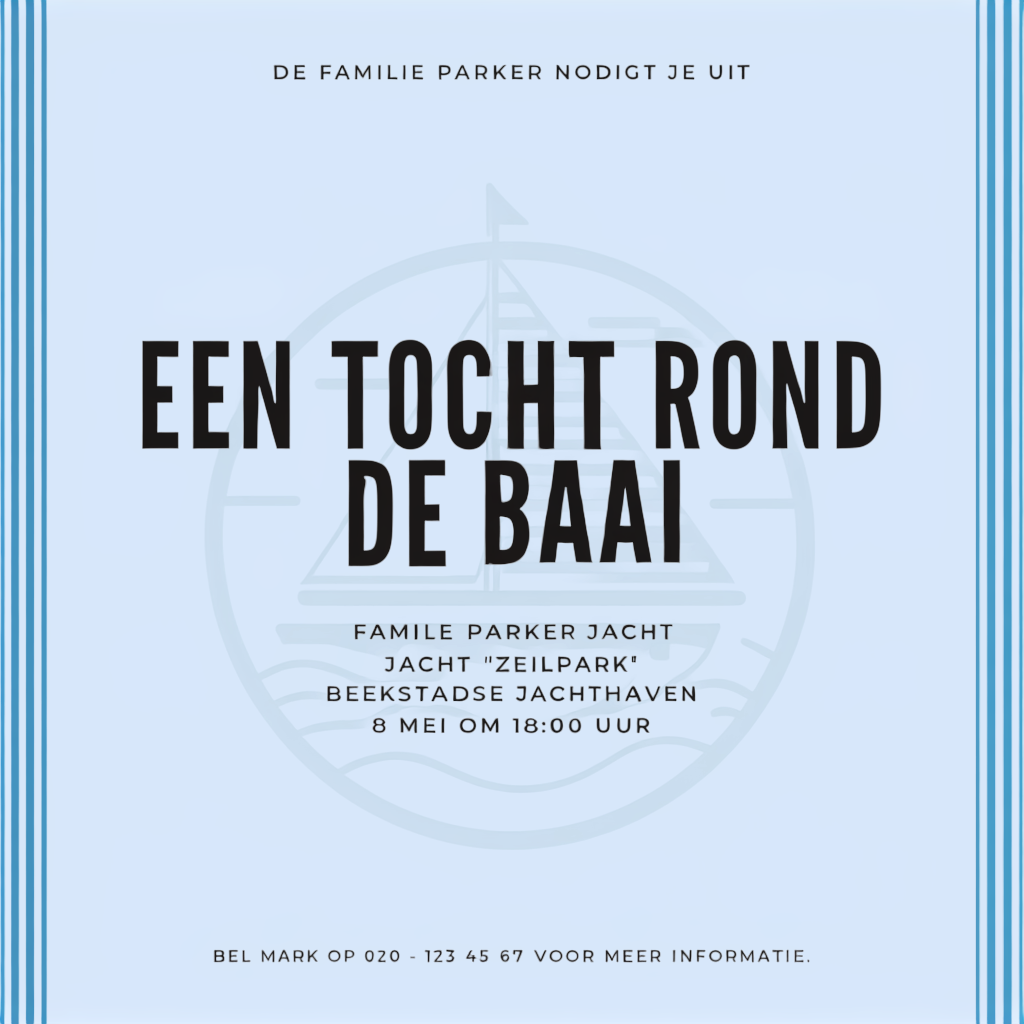}
    \caption{Sailing invitation.}
  \end{subfigure}\\[4pt]
  \begin{subfigure}[b]{0.31\textwidth}
    \includegraphics[width=\linewidth]{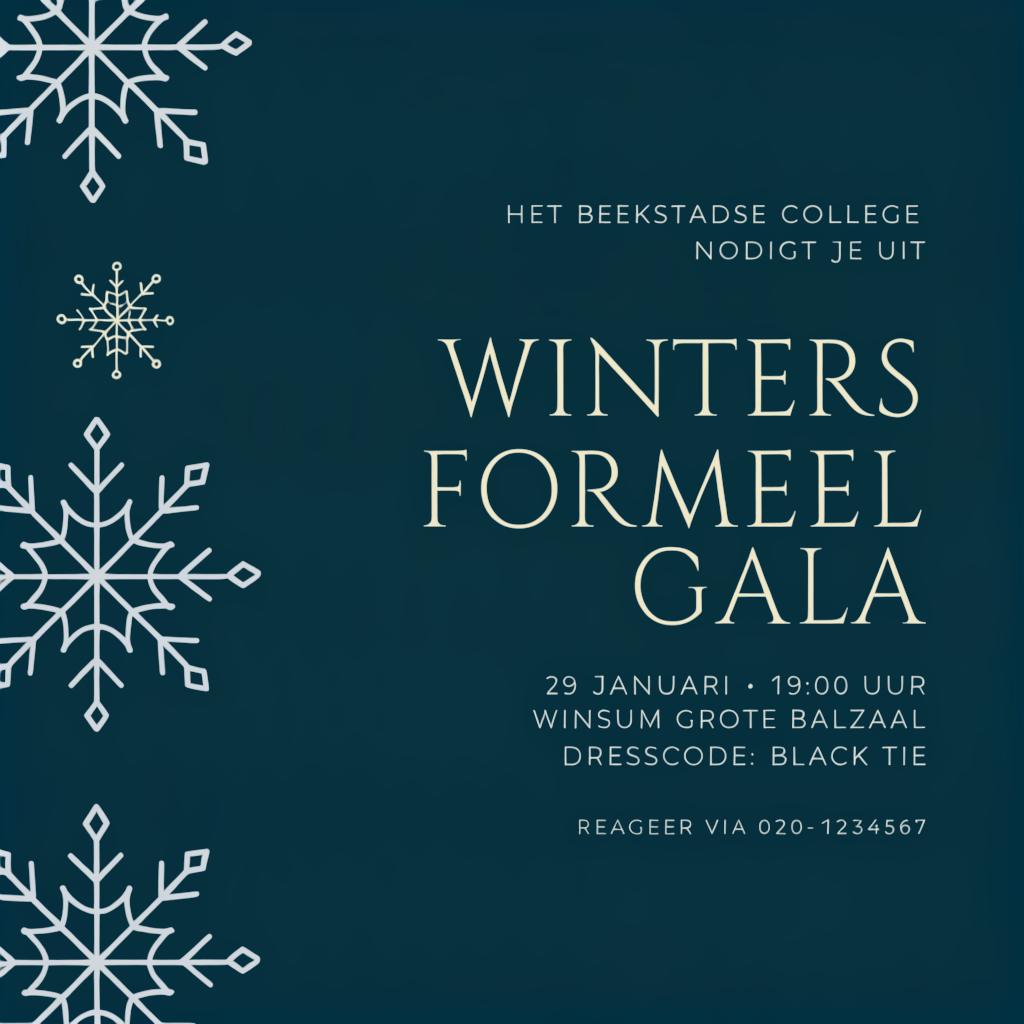}
    \caption{Winter-gala invitation.}
  \end{subfigure}\hfill
  \begin{subfigure}[b]{0.31\textwidth}
    \includegraphics[width=\linewidth]{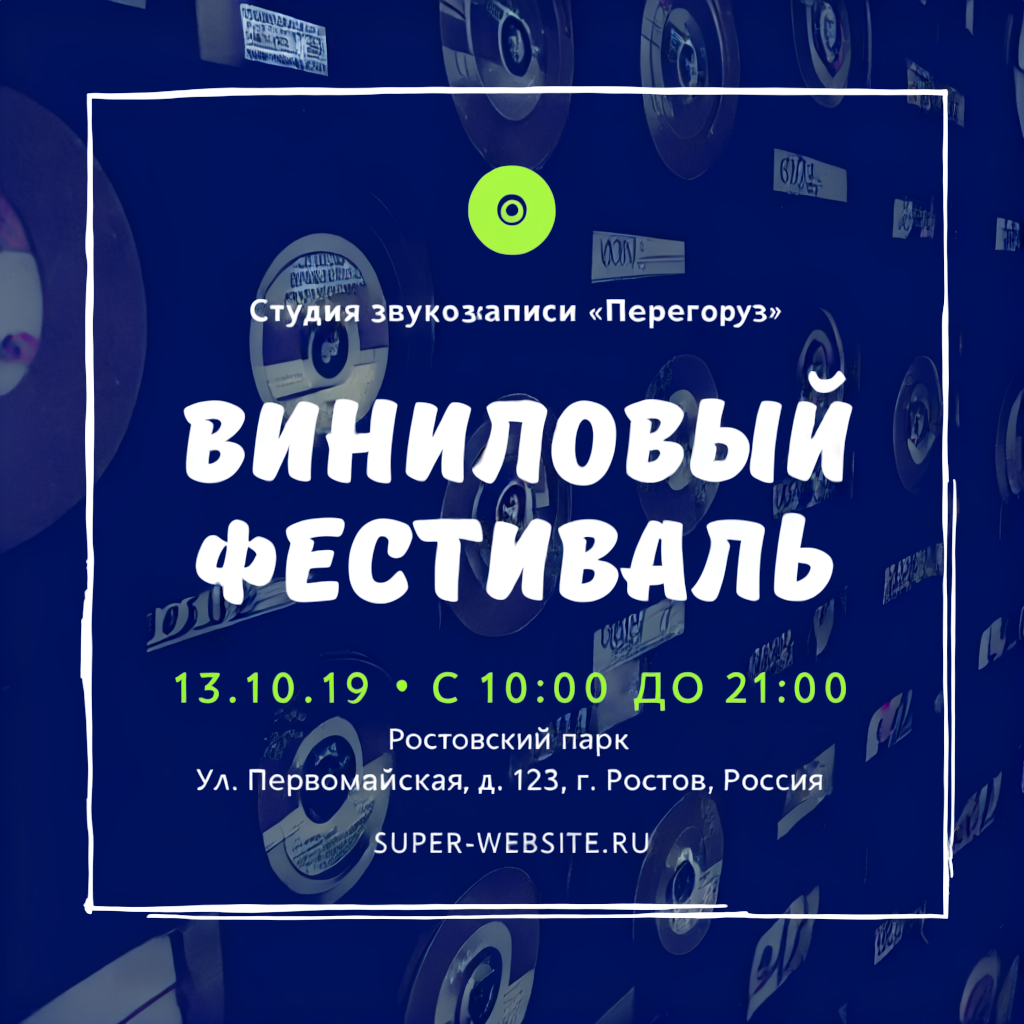}
    \caption{Vinyl-festival poster.}
  \end{subfigure}\hfill
  \begin{subfigure}[b]{0.31\textwidth}
    \includegraphics[width=\linewidth]{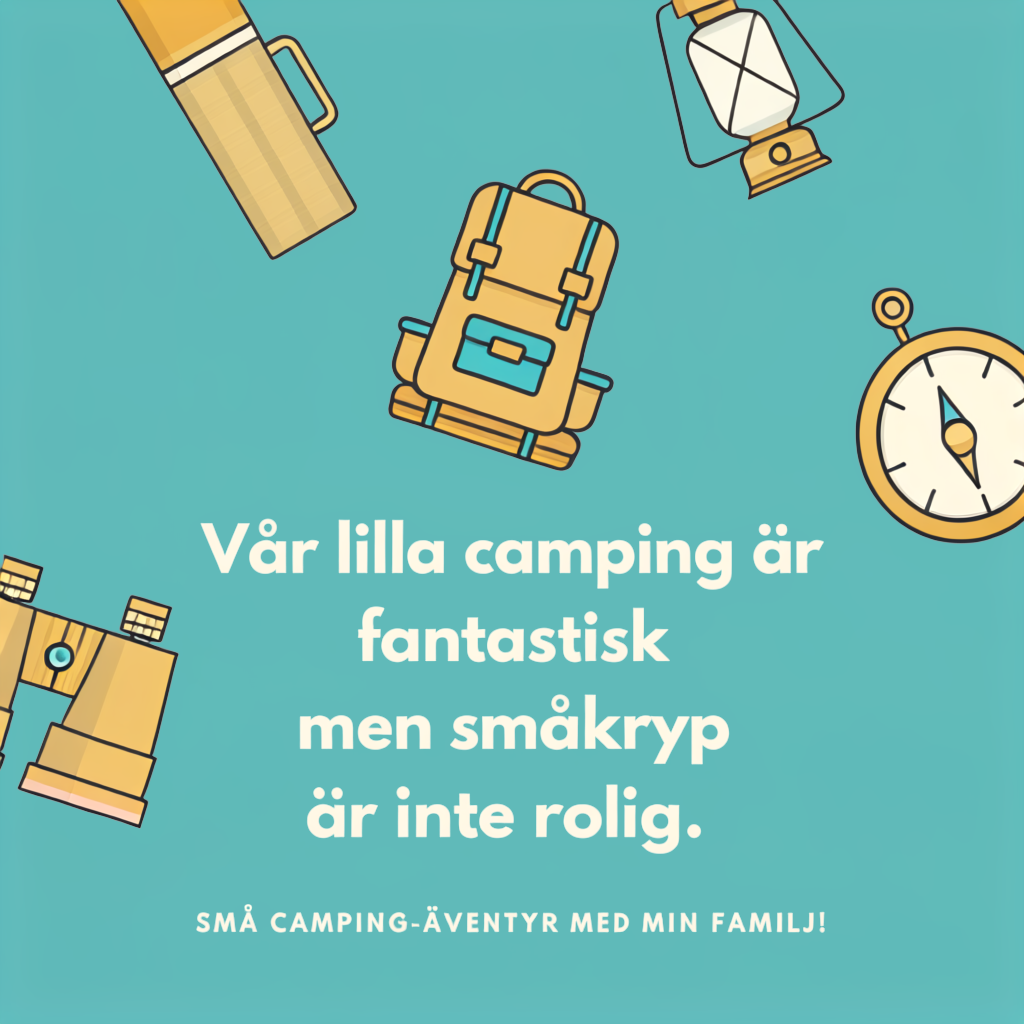}
    \caption{Camping illustration.}
  \end{subfigure}
  \caption{Generated text-to-design composites corresponding to the qualitative layer collages in
  Figure~\ref{fig:text2design-qual} and Figures~\ref{fig:supp-text2design-collages}--\ref{fig:supp-text2design-recipe}.
  They span icon-heavy illustrations, typography-led invitations, and photo-background posters.}
  \label{fig:supp-text2design-composites}
\end{figure*}

\begin{figure*}[tbp]
  \centering
  \includegraphics[width=\textwidth]{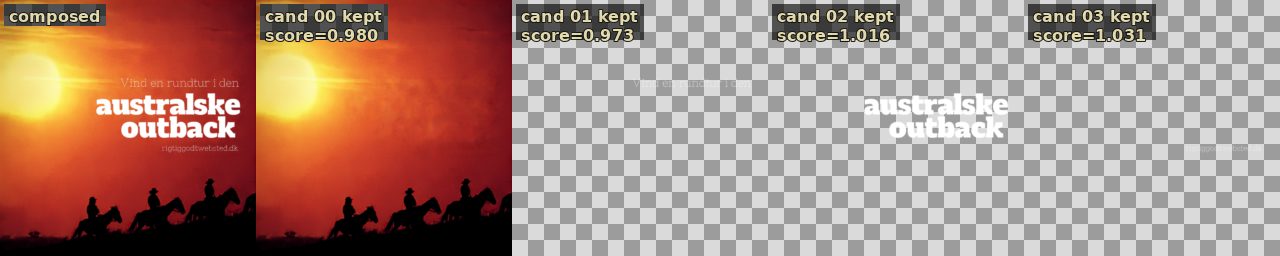}\\[-2pt]
  \includegraphics[width=\textwidth]{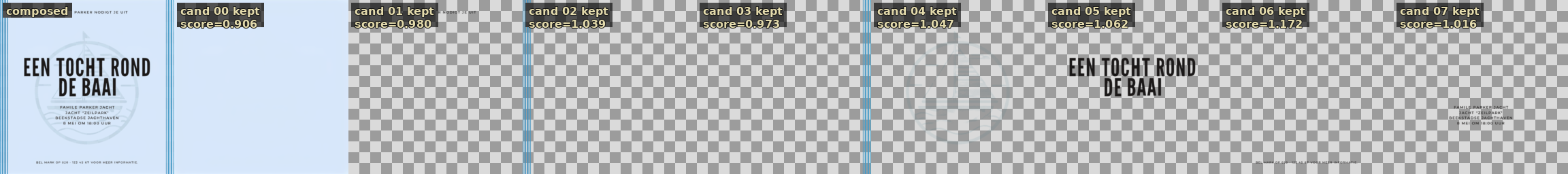}\\[-2pt]
  \includegraphics[width=\textwidth]{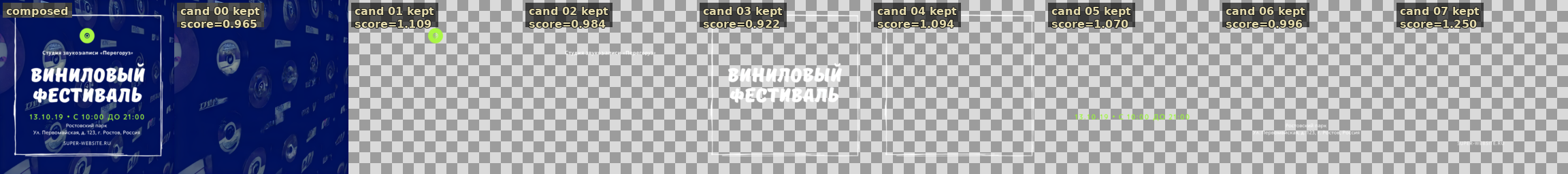}
  \caption{Additional text-to-design layer collages. Top: the outback poster separates a sunset
  background, silhouette, and multiple text components. Middle: the sailing invitation isolates the
  emblem and typography, with several low-visibility text layers. Bottom: the vinyl-festival
  poster illustrates the photo-background failure mode: despite some recovered foreground text, multiple
  layers retain parts of the full background and border.}
  \label{fig:supp-text2design-collages}
\end{figure*}

\begin{figure*}[tbp]
  \centering
  \includegraphics[width=\textwidth]{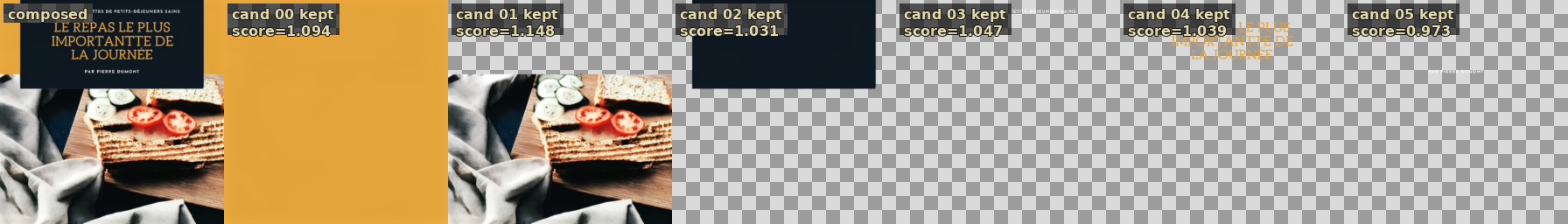}
  \caption{Additional text-to-design candidate collage for a recipe poster. The food photograph is
  retained primarily in one layer while the title and small text are distributed across separate,
  partly faint layers. This case also contains the duplicated-letter rendering artifact discussed
  in Section~\ref{sec:exp-t2d-qual}.}
  \label{fig:supp-text2design-recipe}
\end{figure*}
\subsection{Composition-Preserving Layer Alignment}
\label{sec:supp-alignment}

All methods in Table~\ref{tab:planned-diffusion} are evaluated with this same
composition-preserving alignment, accommodating equivalent layer orders and differing
decomposition granularities.

DTW constrains correspondences to be monotonic in the supplied layer sequences; it does not
establish that a design has a unique stacking order. In particular, layers with disjoint alpha
support commute under source-over compositing, so distinct sequences can render the same design.
Visible occlusions constrain ordering, but non-overlapping elements need not share a unique total
order across predicted and reference decompositions. A monotonic alignment can therefore penalize
an otherwise valid correspondence. Different decomposition granularities introduce a further need
for grouping. We distinguish this evaluation issue from errors in the generated layers themselves.

Composition-preserving alignment searches canonical disjoint groupings of predicted and ground-truth layers, starting
from both original layers and the baseline's initial grouping. Group members are composited in
ascending original-layer order. Visible precedence constraints determine an admissible group order;
we reject cycles and require the full premultiplied RGBA render to be preserved within $10^{-12}$
in FP64 on each side. The search permits up to five merges, retaining at least two groups on an
edited side. For each admissible pair of partitions, a minimum-mean-cost bipartite edge cover
selects correspondences covering every group, with repeated group participation allowed.
Render preservation constrains grouping and stacking within each design; correspondence edges
associate the resulting groups without physically reordering either stack. The alignment cost
is $(\mathrm{RGB\ L1}+1-\mathrm{alpha\ IoU})/2$. We retain the original baseline as an incumbent.
Exhaustion certifies only the reachable bounded canonical search space; timed-out results remain
best-known solutions.

\begin{figure*}[tbp]
  \centering
  \includegraphics[width=\textwidth]{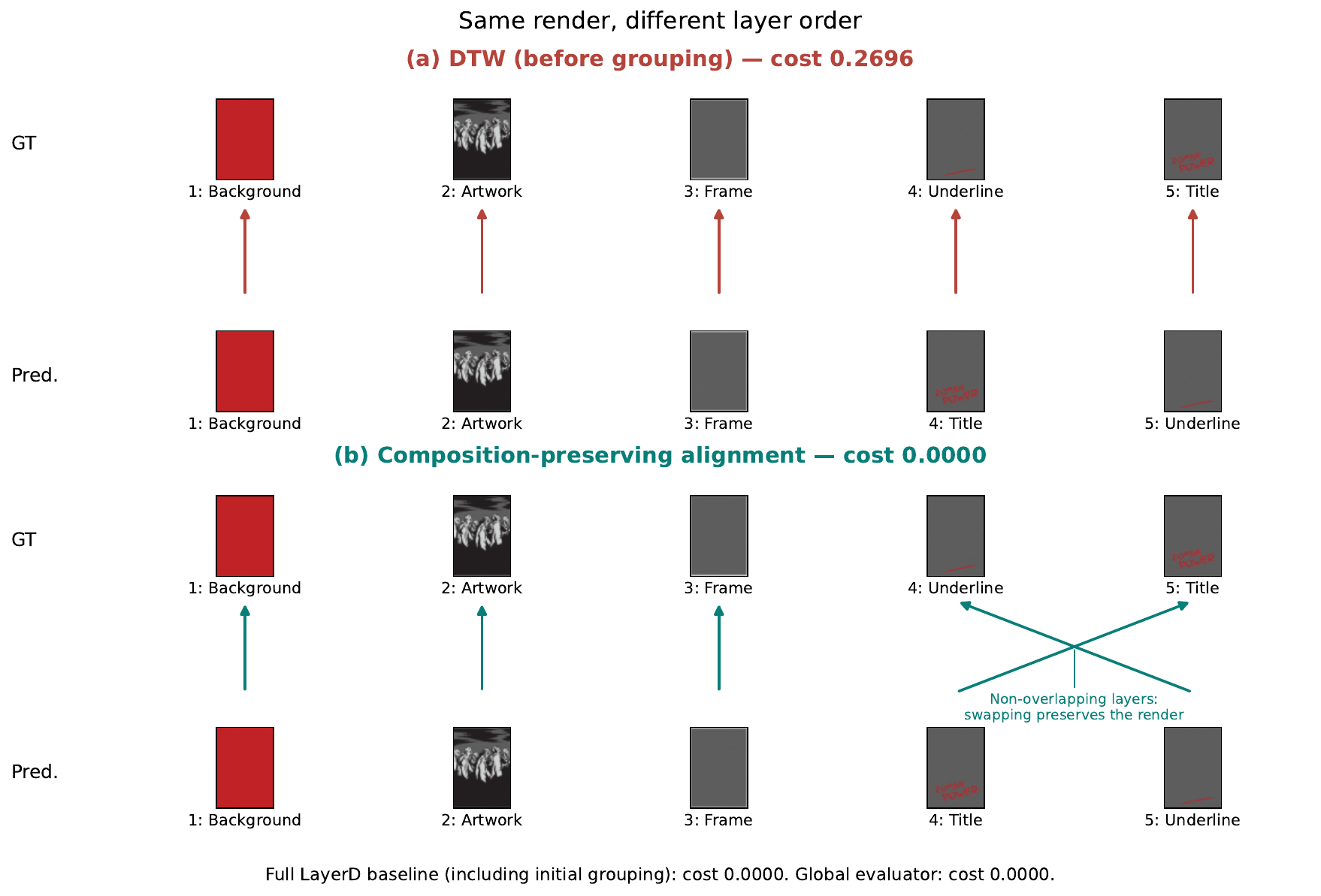}
  \caption{\textbf{Layer correspondence without grouping or merges.}
  Moving the ``Zombie Power'' title from slot 5 to slot 4 preserves the render because
  the title and underline do not overlap. On these ungrouped stacks, monotonic DTW
  mismatches the two layers, whereas composition-preserving correspondence matches them
  exactly with crossing edges and zero merges. This controlled permutation uses observed
  layers, not model-generated predictions.}
  \label{fig:supp-alignment-comparison}
\end{figure*}

With no grouping or merges applied to either stack, the permutation in
Figure~\ref{fig:supp-alignment-comparison} yields joint cost $0.2696$, RGB L1 $0.1391$,
and alpha soft-IoU $0.6000$ under DTW, versus $0$, $0$, and $1$ with unrestricted
correspondence. The full premultiplied RGBA renders are identical.
This isolates the ordering constraint: the complete LayerD baseline also scores zero
at its zero-edit setting, which includes an initial non-overlap grouping.

\end{document}